\documentclass[final,5p,times,twocolumn]{elsarticle}

\usepackage{amsmath, amssymb, amsfonts}
\usepackage{amsthm}

\usepackage{graphicx}
\usepackage{subcaption}   

\usepackage{array}
\usepackage{booktabs}
\usepackage{tabularx}
\usepackage{makecell}
\usepackage{colortbl}
\usepackage{rotating}
\usepackage{multirow}
\usepackage{multicol}

\usepackage[table,xcdraw]{xcolor}
\usepackage{hyperref}

\usepackage{algorithmic}

\usepackage{textcomp}
\usepackage{stfloats}
\usepackage{url}
\usepackage{verbatim}
\usepackage{afterpage}
\usepackage{floatrow}
\usepackage{siunitx}
\usepackage{epstopdf}
\usepackage{enumerate}
\usepackage{natbib}
\usepackage{cleveref}

\usepackage[switch]{lineno}

\journal{Computers and Electronics in Agriculture}
\begin{document}
\definecolor{redbg}{RGB}{250, 0, 0}
\definecolor{yellowbg}{RGB}{250, 250, 0}
\definecolor{bluebg}{RGB}{0, 110, 250}
\newcolumntype{C}{>{\centering\arraybackslash}X}
\begin{frontmatter}
\title{LeafInst - Unified Instance Segmentation Network for Fine-Grained Forestry Leaf Phenotype Analysis: A New UAV based Benchmark}


\affiliation[njfu]{organization={College of Information Science and Technology \& Artificial Intelligence, Nanjing Forestry University},
	city={Nanjing},
	postcode={210037},
	state={Jiangsu},
	country={China}
}

\affiliation[vt]{organization={Department of Geography, College of Natural Resources and Environment, Virginia Tech},
	city={Blacksburg},
	state={VA},
	postcode={24061},
	country={USA}
}

\affiliation[coi]{organization={Co-Innovation Center for Sustainable Forestry in Southern China, Nanjing Forestry University},
	city={Nanjing},
	postcode={210037},
	state={Jiangsu},
	country={China}
}

\affiliation[szu]{organization={School of Architecture \& Urban Planning, Shenzhen University},
	city={Shenzhen},
	postcode={518060},
	state={Guangdong},
	country={China}
}

\affiliation[hfut]{organization={School of Software, Hefei University of Technology},
	city={Hefei},
	postcode={230601},
	state={Anhui},
	country={China}
}


\author[njfu,vt]{Taige Luo\fnref{fn1}}
\author[njfu,vt]{Junru Xie\fnref{fn1}}
\author[njfu]{Chenyang Fan}
\author[hfut]{Bingrong Liu}
\author[szu]{Ruisheng Wang}
\author[vt]{Yang Shao}
\author[njfu]{Sheng Xu\corref{cor1}}
\author[coi]{Lin Cao}

\cortext[cor1]{Corresponding author: Sheng Xu (xusheng@njfu.edu.cn)}
\fntext[fn1]{These authors contributed equally to this work.}

\begin{abstract}
Intelligent forest tree breeding has spurred significant advancements in plant phenotyping research. However, the existing studies predominantly concentrate on agricultural scenarios involving large-leaf crops, leaving a notable gap in the fine-grained research on the leaves of sapling trees in open-field environments. Moreover, leaves in natural settings present potential challenges due to variations in scale, brightness, and shape. To tackle these critical issues, this study employed unmanned aerial vehicles (UAVs) to survey field-grown saplings and collect 1,202 branches, encompassing a total of 19,876 leaf instances. Leveraging this data, we introduce and publicly release an open-source UAV RGB remote sensing dataset, named Poplar-leaf. Significantly, this dataset represents the first pixel-level annotated instance segmentation dataset specifically tailored for forestry leaves in open-field scenes. Regarding accurate segmentation, a novel network architecture, LeafInst, is proposed. For scale changes, the Asymptotic Feature Pyramid Network (AFPN) is integrated into the Neck component, which effectively enhances progressive multi-scale visual perception. Additionally, a novel Dynamic Asymmetric Spatial Perception (DASP) module is designed to optimize feature extraction for irregularly shaped leaves. Based on it, we propose a dual-residual Dynamic Anomalous Regression Head (DARH) block to enhance the detection feature fusion. The proposed Top-down Concatenation–decoder Feature Fusion (TCFU) fuses the output with the bounding box regression branch, effectively mitigating redundant feature representations. Quantitative experiments demonstrate that, on the Poplar-leaf dataset, LeafInst achieves a 7.1\% improvement in segmentation mAP (68.4) compared with YOLOv11, and a 6.5\% gain over the transformer-decoder-based MaskDINO. Furthermore, on the public large-scale agricultural benchmark dataset PhenoBench, LeafInst outperforms MaskDINO by 3.4\% in box mAP, reaching 52.7 mAP. Qualitative experiments show that LeafInst outperforms the large visual models in zero-shot scenes. Finally, we present two downstream applications to highlight the real-world utility of the proposed model, while also supporting its robustness across diverse scenarios: robust leaf segmentation in complex scenes, and a novel high-throughput Leaf Growth Condition Indicator (LGCI) for quantitative assessment of leaf phenotypic development (from 1850 leaf instances). \\
\textbf{Keywords:} Deep Learning, Computer Vision, UAV remote sensing, Instance segmentation, Leaf phenotype analysis, Intelligent forestry
\end{abstract}
\end{frontmatter}

\section{Introduction}

With the advancement of smart forestry, research combining "intelligent" models with practical applications has been growing rapidly. Current deep learning-based studies have covered numerous domains, including wheat ear phenotyping identification \cite{02FANwheat,13wheat}, forest tree species recognition \cite{04urbantrees,12Cao}, fine-grained classification of forest land use \cite{03Flanduse,07vegem,10forestmo}, and crop disease/pest detection \cite{01FANdisease,05diseases,06pest}. These studies have demonstrated significant practical value. As the most critical component of forests, trees play a vital role in understanding forest dynamics. Forestry breeding represents a significant silvicultural challenge that relies on precise plant phenotyping analysis to select superior varieties with optimal growth characteristics, luxuriant foliage, and adequate water content \cite{08breeding,11cropbreedng,14breeding}. Conventional breeding remains an effective approach for forest tree improvement internationally. Leading forestry nations should place greater emphasis on phenotypic and genetic evaluation of germplasm resources, along with the exploration and utilization of unique genetic materials, to establish comprehensive core collections. Long-term breeding programs have entered a phase of accelerated development, where continuous genetic gains in improved varieties are being achieved through techniques such as hybrid breeding and multiple-trait integration breeding. However, systematic genetic evaluation of germplasm resources and core collection establishment have not been conducted for most tree species. 
\begin{figure*}[t]
	\centering
	\includegraphics[width=1\textwidth]{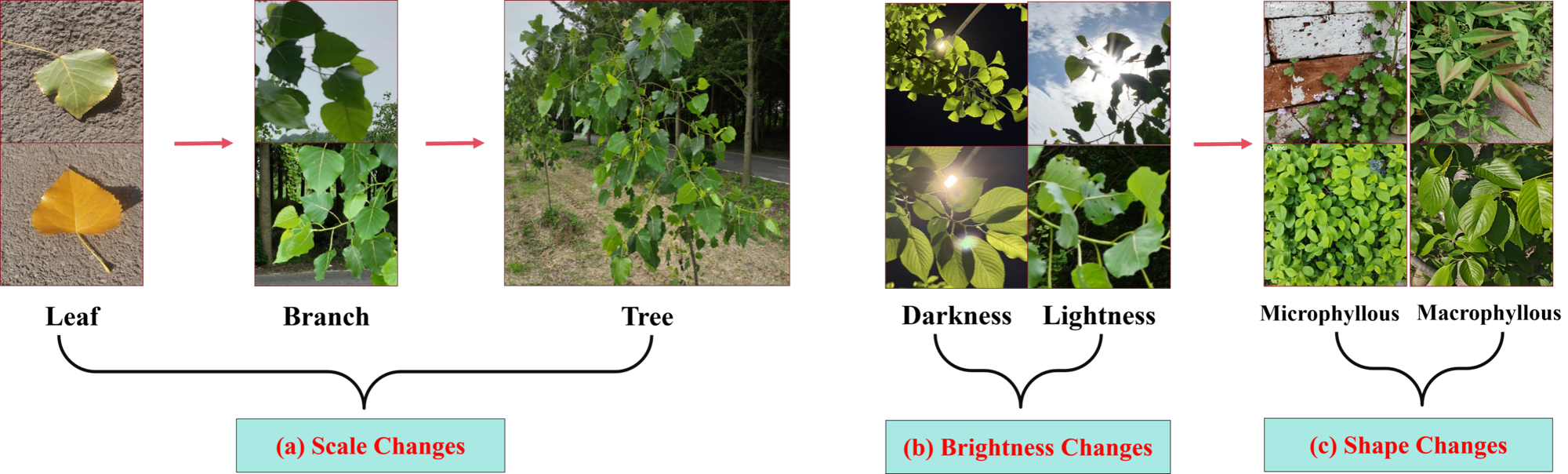}
	\caption{Challenges in forestry leaf phenotyping. (a) Scale changes, (b) Brightness changes, (c) Shape changes.}
	\label{fig:1}
\end{figure*}
Key technologies such as parental selection, early-stage selection, and stable high-yield seed orchards remain unbroken. The traditional manual evaluation system relies on field inspections, which are time-consuming, labor-intensive, subjective, and potentially damaging to parent trees. This system fails to establish intelligent standardized evaluation criteria and frameworks, resulting in current breeding practices that cannot meet growing consumer demands. In terms of production capacity, taking China as an example, official statistics from 2016 indicate that while the country possesses 4.68 billion mu (approximately 312 million hectares) of forest land, it faces severe timber shortages. The total timber consumption reached 522 million cubic meters, with projections suggesting an annual supply-demand gap exceeding 450 million cubic meters post-2025. This implies an annual transition of 50 million cubic meters of timber production from natural forests to plantations. Currently, most seedlings transplanted to plantation areas are immature, and traditional manual analysis methods demonstrate poor adaptability. 
\begin{figure*}[h]
	\centering
	\includegraphics[width=0.9\textwidth]{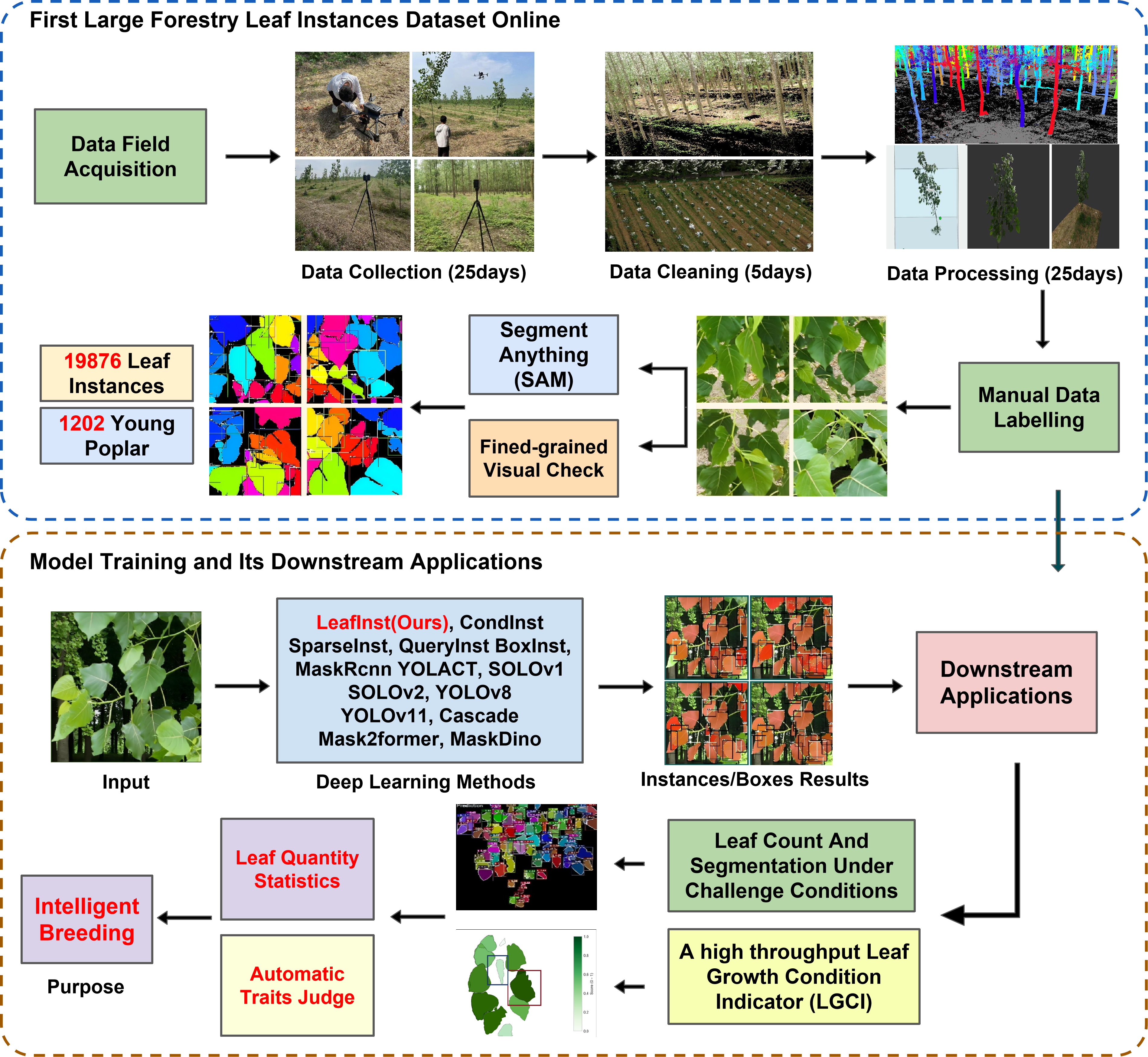}
	\caption{The workflow and downstream applications of the proposed framework LeafInst.}
	\label{fig:2}
\end{figure*}
Conventional breeding programs primarily emphasize "fast growth, stress resistance, and wood quality" while neglecting the carbon sequestration capacity of individual tree leaves. A tree's carbon sequestration potential is directly determined by its biomass accumulation rate, which correlates closely with leaf phenotypic traits (e.g., Leaf Area Index (LAI)\cite{15LAI}, Specific Leaf Area (SLA) \cite{16SLA}, trunk morphology (DBH) \cite{17DBH}, height increment \cite{18HI}, and root architecture \cite{20root,59ye1}). "Early-stage selection" has emerged as a core challenge in forest tree improvement. As a fundamental indicator of vegetation growth, leaf-derived parameters can effectively evaluate developmental potential in trees \cite{19leaf}. \par

Individual Tree Segmentation (ITS) has emerged as a research focus in recent years. However, forest tree breeding requires finer-grained phenotypic parameters, driving the trend toward more "granular" research subjects. To assist agricultural/forestry practitioners and ecologists in extracting fine-grained tree structures, analyzing plant phenotypes, and improving superior variety selection, researchers have conducted experiments on tree branches \cite{09branch,25b2,26b3}, crown width \cite{21cw1,22cw2,23cw3}, and DBH \cite{24dbh1,27dbh2}. These studies significantly contribute to understanding plant health status and predicting growth trends, enabling the selection of elite varieties at early seedling stages for cultivating high-yield specimens. Nevertheless, few forestry studies address the most critical phenotypic parameter - leaves. Current leaf-level research primarily focuses on visibly diseased/pest-infected leaves \cite{28p1,29p2} and agricultural crop leaves \cite{30c1,31c2,32c3}. This study represents the first instance-based investigation of forest tree leaves, establishing a fundamental milestone for this field. Building upon agricultural "crop phenomics platform" experiences, this framework can be extended to deploy fixed cameras in plantation test areas to analyze leaf growth rates through video streams, develop dynamic genetic gain evaluation models, and open-source the dataset on global platforms. These efforts will provide data support for worldwide forest ecological breeding initiatives. \par

Compared to agricultural crops, tree leaves present distinct challenges: dense distribution, broad leaf morphology, and close-range sampling. In addition, agricultural leaves often present orthorectified images, which reduce the angles transformation. However, the height and density of trees expose them to environmental variables, leading to uneven illumination, high occlusion density, and morphological deformations. Overall, forestry leaf UAV images present three major challenges: 1. Scale changes (as the drone lens expands and contracts). Scale variations arise from changes in camera configurations and task requirements, leading UAV imagery to focus on different spatial levels, including individual leaves, branches, and entire trees. 2. Brightness Changes (as sampling time changes). Brightness variations arise from changes in sampling time: insufficient illumination at night and varying illumination angles and intensities during daytime introduce diverse shadow patterns, posing significant challenges for accurate instance extraction. 3. Shape changes (varying with the sampling object), due to the species diversity of trees, we divide leaf types into microphyllous and macrophyllous . We illustrate these challenges in Figure \ref{fig:1}, which significantly increases recognition complexity.\par 

To address the scarcity of forestry leaf data, we conducted UAV circumnavigation flights around individual poplars in Dongtai Plantation, capturing multi-angle, variable-lighting leaf samples. Assisted by AI foundation model Segment Anything(SAM) \cite{33SAM} with forestry expert annotation, we developed the Poplar-leaf dataset - the first pixel-level annotated, high-quality open-source dataset specifically designed for juvenile poplar leaves instances segmentation. For this, our research team spent nearly two months screening out plants with good traits and growth periods. Our proposed LeafInst model demonstrates exceptional zero-shot transfer capability to other agricultural/forestry leaf analysis tasks by training with Poplar-leaf. The proposed workflow is detailed in Figure \ref{fig:2}. Additionally, we propose two practically valuable downstream applications. Firstly, we use LeafInst to finish leaf instances segmentation tasks under different challenge scenarios (Scale Changes, Brightness Changes, Shape Changes). Secondly, we propose a novel high-throughput indicator, LGCI. This indicator is used to quantify the growth and development of leaves, implying leaf shape and band information. It can be used to infer the growth trend of leaves and conduct breeding selection. These applications directly address critical bottlenecks in forest tree breeding. Contributions are as follows:
\begin{itemize}
	\item We present an open-source Poplar-leaf dataset. This establishes a much-needed benchmark for forest leaf segmentation, which was previously absent in the remote sensing study, serving as a foundation for future ecological studies. It collected 1,202 young poplar branches via UAV field sampling, comprising 19,876 leaf instances. Additionally, this study provides approximately 40.8 GB of poplar sapling instances covering 20,000 square meters for the extended research. 
	
	\item A high-precision segmentation framework, LeafInst, is proposed to deal with leaf instance segmentation at the scale of individual leaves, individual branches, and individual trees from different forestry scenarios. We propose a novel unified DARH head built in DASP to enhance the detection accuracy for three challenges. Additionally, we re-explore existing feature fusion strategy and propose a simple but useful TCFU strategy to solve the problem of low-level feature redundancy. Compared to existing SOTA algorithms, the proposed network demonstrate its superiority by quantitative and qualitative comparison.
	
	\item We proposed a high-throughput leaf growth refinement expression indicator LGCI for quantitative assessment of leaf phenotypic development based on predicted instances (over 1850 instances, with most $R^{2}$ on average of 0.9 for secondary indicators).
\end{itemize} \par
Experiments show that the proposed Poplar-leaf dataset bridges the gap between instance segmentation and Leaf phenotype analysis in the open-field scenes. LeafInst has surpassed the existing models and demonstrated strong robustness in subsequent application challenges, which is demonstrated by downstream applications for forestry breeding, especially in the growth analysis of leaves with high throughput.

\section{Related Work}
\subsection{Plant Phenotype Research Based on the Forestry Remote Sensing Technologies}
With the advancement of remote sensing technologies, an increasing number of studies have focused on forestry remote sensing. In terms of spatial scale, existing research can be broadly categorized into large-scale vegetation inversion at coarse spatial resolution and fine-scale vegetation inversion at high spatial resolution. Traditional methods employ remote sensing imagery for large-scale vegetation mapping and inversion. For example, 
Lukas et al.\cite{34lukas} utilized the radiative transfer model (RTM) to simulate the directional reflectance of winter wheat by combining the PROSPECT-D leaf RTM and 4SAIL canopy RTM. The RTM inputs include physical characteristics of leaves and canopies, such as chlorophyll content and leaf angle. This research successfully monitored wheat growth and phenological stages at a landscape scale. Ren et al.\cite{35ren} explored annual wheat growth patterns using time-series MODIS NDVI 8-day composite data, establishing a foundation for understanding the crop's intensity dynamics from 2001 to 2003. Lukas et al.\cite{34lukas} suggested that future research should focus more on scaling leaf features to canopy levels and investigating finer-grained phenological stages.\par
Currently, fine-grained individual tree modeling methods can be categorized into two types based on data availability: approaches using 2D optical images and those employing 3D LiDAR point clouds. These two branches have significantly advanced the field of individual tree modeling and contributed substantially to understanding ecological indicators. In the past, due to hardware limitations, high-resolution optical images were difficult to obtain, and LiDAR point cloud-based methods primarily relied on traditional machine learning. In the domain of individual tree segmentation, these methods have made commendable attempts to analyze tree skeleton structures and acquire vegetation phenotypic parameters. For instance, Xu et al.\cite{36stem} applied an optimized point clustering method and achieved promising results for stem extraction. While Li et al.\cite{37Li} proposed a novel individual tree detection direction based on multichannel representation. However, traditional machine learning-based approaches require substantial time for data collection and struggle to meet real-time monitoring automatically. With the development of deep learning and GPU hardware, more vision models have emerged. Specifically, deep learning has quickly dominated research due to its superior performance and inference speed \cite{40ITS1,41ITS2}. These studies typically focus on modeling complete tree structures, while research on more refined local features—such as branches \cite{38b}, flower buds \cite{39fl}, and fruits remains scarce \cite{42fr}, especially leaves. However, these key components of trees are very important for breeding. Consequently, some researchers have begun using high-resolution and more accessible optical images for vegetation phenotyping, achieving results comparable to point clouds at lower costs and with faster inference speeds. For example, Fan et al.\cite{02FANwheat} employed a few-shot deep learning approach to complete research on wheat head counting, advancing intelligent crop yield estimation. Cao et al.\cite{12Cao} successfully implemented cross-scale plantation canopy extraction using an optimized YOLO model.\par
Nevertheless, most existing studies focus on agricultural leaves. Some studies, such as that by Barreto et al.\cite{73al1}., investigated leaf phenotyping through instance segmentation across four sugar beet plots. However, due to limitations in imaging viewpoints, the acquired images are presented as orthorectified remote sensing imagery, which makes it difficult to capture the diverse and deformable postures of forestry leaves. In addition, Weyler et al.\cite{74al2}. proposed a novel convolution-based neural network that achieved successful leaf segmentation. Nevertheless, their study was still restricted to sugar beet and weed scenarios and did not further analyze phenotypic parameters. As a result, these methods are difficult to directly transfer to forestry applications. \par
Compared to agricultural leaf instances, tree leaves are denser, unevenly illuminated, and often exhibit irregular shapes due to greater exposure to wind at higher elevations. Moreover, trees have longer growth cycles, making phenotypic data monitoring at each stage important for breeding. The extreme irregularity of leaf morphology increases annotation costs, leading most current studies to rely on mature indoor tree samples \cite{43le1,44le2}. However, these samples fail to adequately reflect environmental challenges in natural settings and are too limited in quantity to support multi-species leaf transfer learning. Given the absence of open-source leaf instance datasets in this field, we address this critical research gap by providing a finely annotated dataset of young poplar leaves in natural environments.

\subsection{Instances Segmentation Models Based on the Deep Learning}

The development of deep learning has significantly advanced computer vision and enabled a variety of downstream applications. Instance segmentation holds significant application value. It combines the characteristics of object detection and semantic segmentation, requiring not only the classification of each pixels but also the discrimination of different instances within the same category. It has demonstrated application prospects in fields such as autonomous driving \cite{45d1,45d2}, medical image analysis \cite{46m1,47m2}, face recognition \cite{48f1}, and remote sensing imagery \cite{50r1,60ye2}. \par

The development of instance segmentation methods has gone through several important stages: from early traditional image processing methods to deep learning-based approaches, and then to end-to-end solutions. Current mainstream methods can be divided into the following categories: two-stage pre-detection instances segmentation, single-stage mask-based instances segmentation, and transformer-based methods. The core difference between two-stage and single-stage networks lies in whether explicit object detection is required and whether detection information is utilized in downstream segmentation.\par
\begin{figure*}[t]
	\centering
	\includegraphics[width=0.9\textwidth]{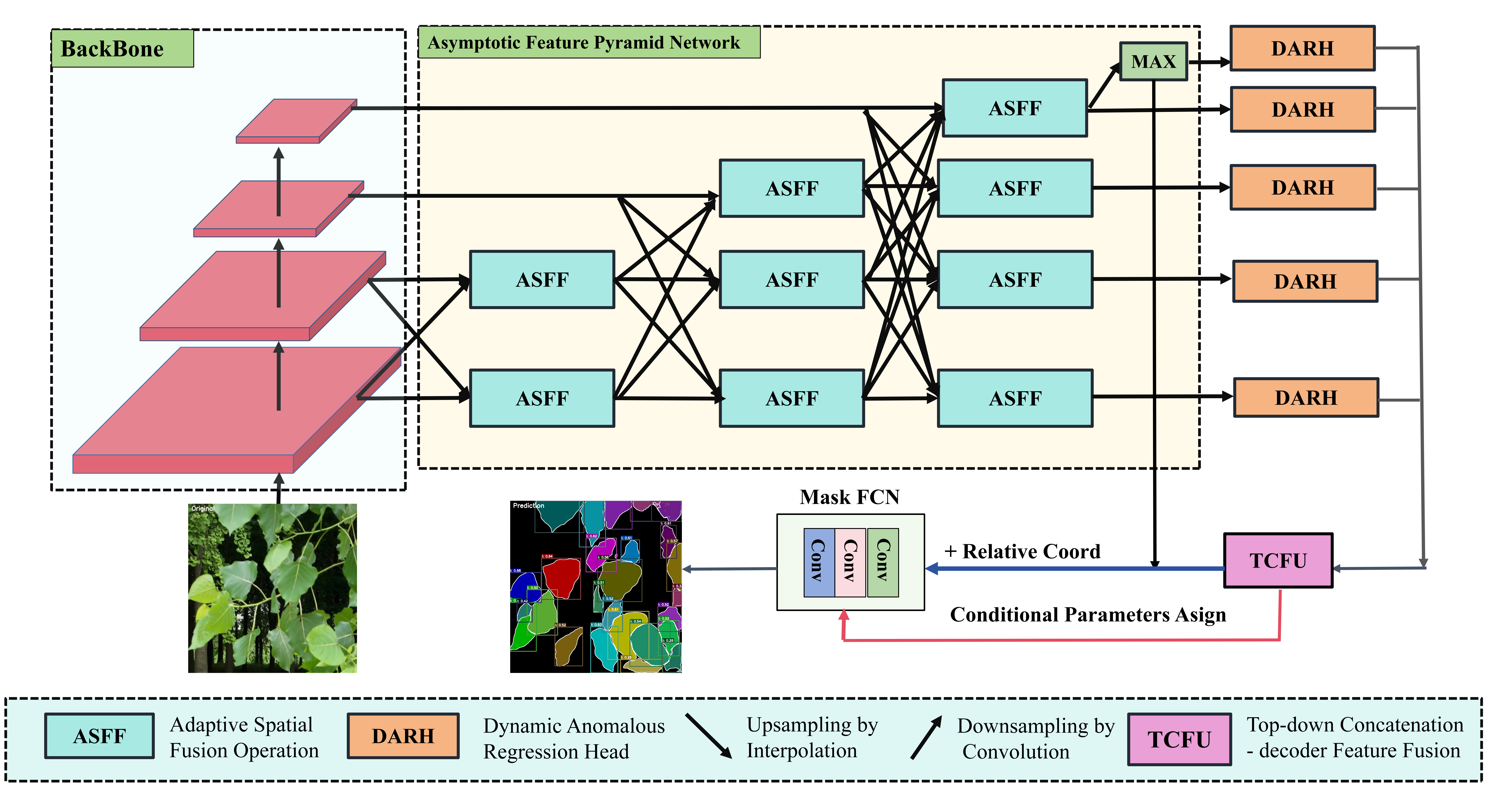}
	\caption{Overview of the main structure of the proposed LeafInst.}
	\label{fig:3}
\end{figure*}
\begin{figure*}[t]
	\centering
	\includegraphics[width=0.8\textwidth]{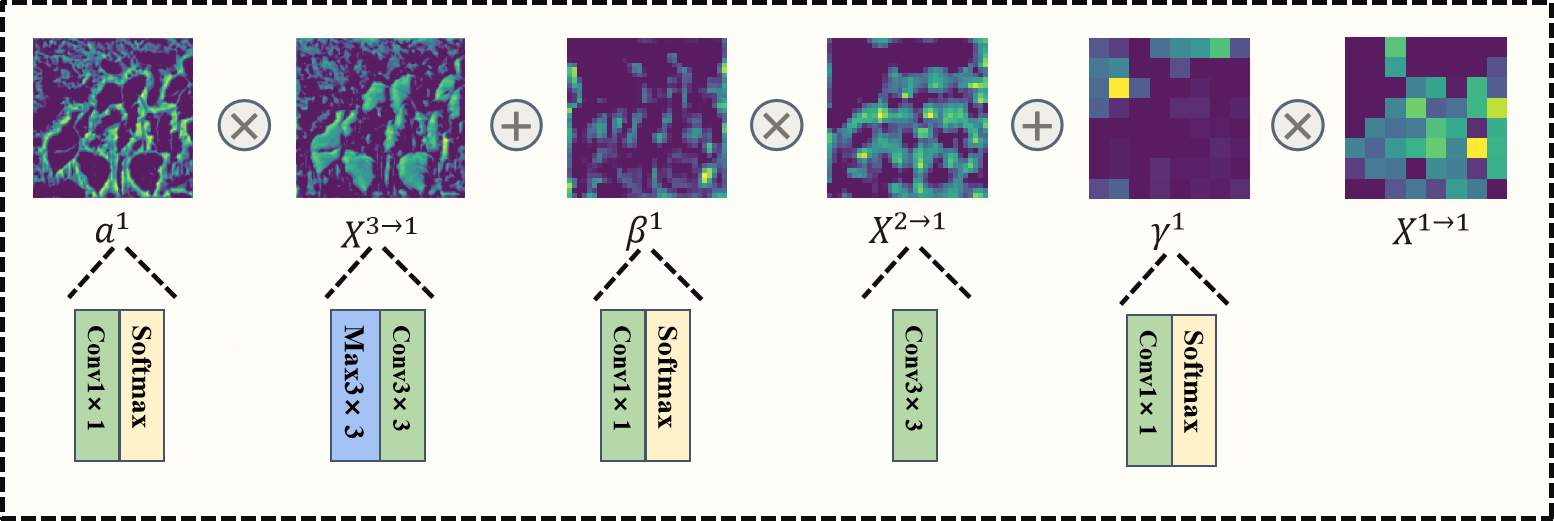}
	\caption{Adaptive spatial fusion operation (ASFF) built in AFPN.}
	\label{fig:4}
\end{figure*}
Early on, Faster-RCNN proposed by Ren et al.\cite{51fas} established a classic paradigm for object detection algorithms, whose core involves matching positive and negative samples through pixels-level anchor boxes and screening potential target boxes via Non-Maximum Suppression (NMS). Subsequently, He et al.\cite{52mask} extended this framework by introducing a segmentation branch built upon the detected bounding boxes. Using ROIAlign to precisely align Regions of Interest (ROIs) with the backbone feature maps, the network predicts instance-level masks, forming the classic two-stage Mask R-CNN architecture. This network is still applied in various baseline tasks and has achieved excellent results. Due to the fixed training threshold of this network, Cai et al.\cite{53cas} proposed an optimized Cascade Mask R-CNN algorithm, which trains multiple cascaded detectors through a preset series of increasing IoU thresholds to obtain more accurate detection and segmentation results. Additionally, BoxInst effectively combines the relationship between bounding boxes and instance masks \cite{54box}. They present a high-performance method that can achieve mask-level instance segmentation with only bounding-box annotations for training.\par
In recent years, due to their complex and interdependent structures, detection-based segmentation algorithms require more time for inference and training, leading to rapid developments in single-stage networks. The YOLACT architecture proposed by Daniel et al.\cite{55yol} divides detection and segmentation into two parallel processes. By generating k prototype masks in the segmentation branch and k mask coefficients in the detection branch for linear combination, it achieves fast real-time segmentation while maintaining acceptable accuracy. SparseInst innovatively introduces a sparse instance representation approach \cite{56spa}. By utilizing sparse points to encode instances, the method empowers the model to directly predict these sparse points from images for instance reconstruction. This design effectively mitigates the substantial redundant computations inherent in conventional instance segmentation techniques. Tian et al.\cite{57con} presents a fully-convolutional architecture. It leverages a set of dynamic convolution kernels to predict the convolution parameters of the Mask branch. This approach attains high accuracy with an limited number of pre-set mask branch parameters which are only 169. Moreover, it effectively strikes a balance between real-time performance and accuracy. Transformer-based methods, such as Mask2former \cite{58mask2}, adopt the Cross Attention design concept from the DETR object detection architecture. They abstract the instance segmentation process into a mask classification problem and accomplish instance segmentation by predicting binary masks corresponding to different categories. However, Transformer models come with certain limitations. They demand a vast amount of data for training. Real-time inference and detection are challenging to implement because their computational complexity is restricted by multi-head attention mechanisms, which is usually exponential.\par
In summary, while the aforementioned methods all possess their own merits, there remains a dearth of specific enhancements tailored to the formidable leaf scenarios characterized by significant overlaps, inadequate illumination, and a wide variety of morphologies. This paper puts forward an instance segmentation algorithm named LeafInst, which is grounded in the optimization of the anchor-free CondInst network\cite{57con}. This algorithm effectively tackles the aforementioned problem. \par
\section{Methodology}
LeafInst is an algorithm framework specifically designed for leaf instance segmentation as shown in Figure \ref{fig:3}. Next, we will explain the component modules and specific functions of the model in the order of feature flow input, starting from the overall view and then delving into the details. In the Neck part of the model, we integrate the Asymptotic Feature Pyramid Network technology for multi-scale feature extraction. This meets the stringent requirements of leaf extraction for skeletal features and compensates for the loss of the receptive field of features at different resolutions. The details will be elaborated in Section 3.1. At the Head end, we propose a novel Dynamic Anomalous Regression Head (DARH) module. It incorporates Dynamic Asymmetric Spatial Perception (DASP) within four different types of feature extraction branches and dual-residual feature fusion strategy. By using four different shapes of convolution operations, it can capture the features of deformed leaves and successfully helps the model recognize multi-variant deformation feature maps. The details will be elaborated in Sections 3.2 and 3.3. After extracting these original features, we input them into a unified multi-scale feature fusion module, namely Top-down Concatenation-decoder Feature Fusion (TCFU) replacing original Top-down Pixel-wise Addition method. This method effectively avoids redundant calculations of low - level features, prevents the problem of feature information weight loss caused by repeated weighting, and improves the feature recombination in the region of interest. The details will be elaborated in Section 3.5.
\begin{figure*}[t]
	\centering
	\includegraphics[width=1\textwidth]{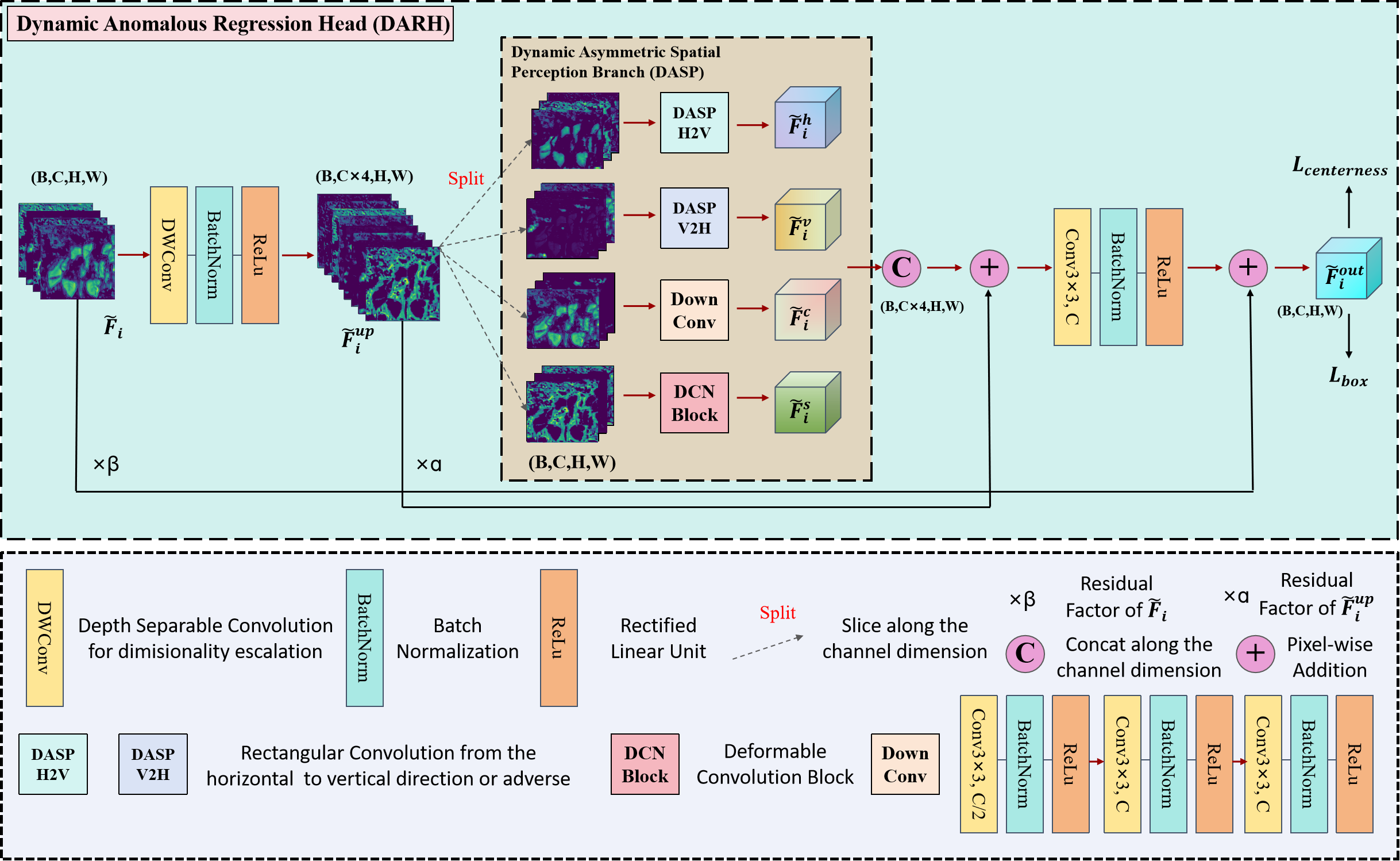}
	\caption{The workflow of dynamic anomalous regression head.}
	\label{fig:5}
\end{figure*}
\begin{figure*}[t]
	\centering
	\includegraphics[width=0.7\textwidth]{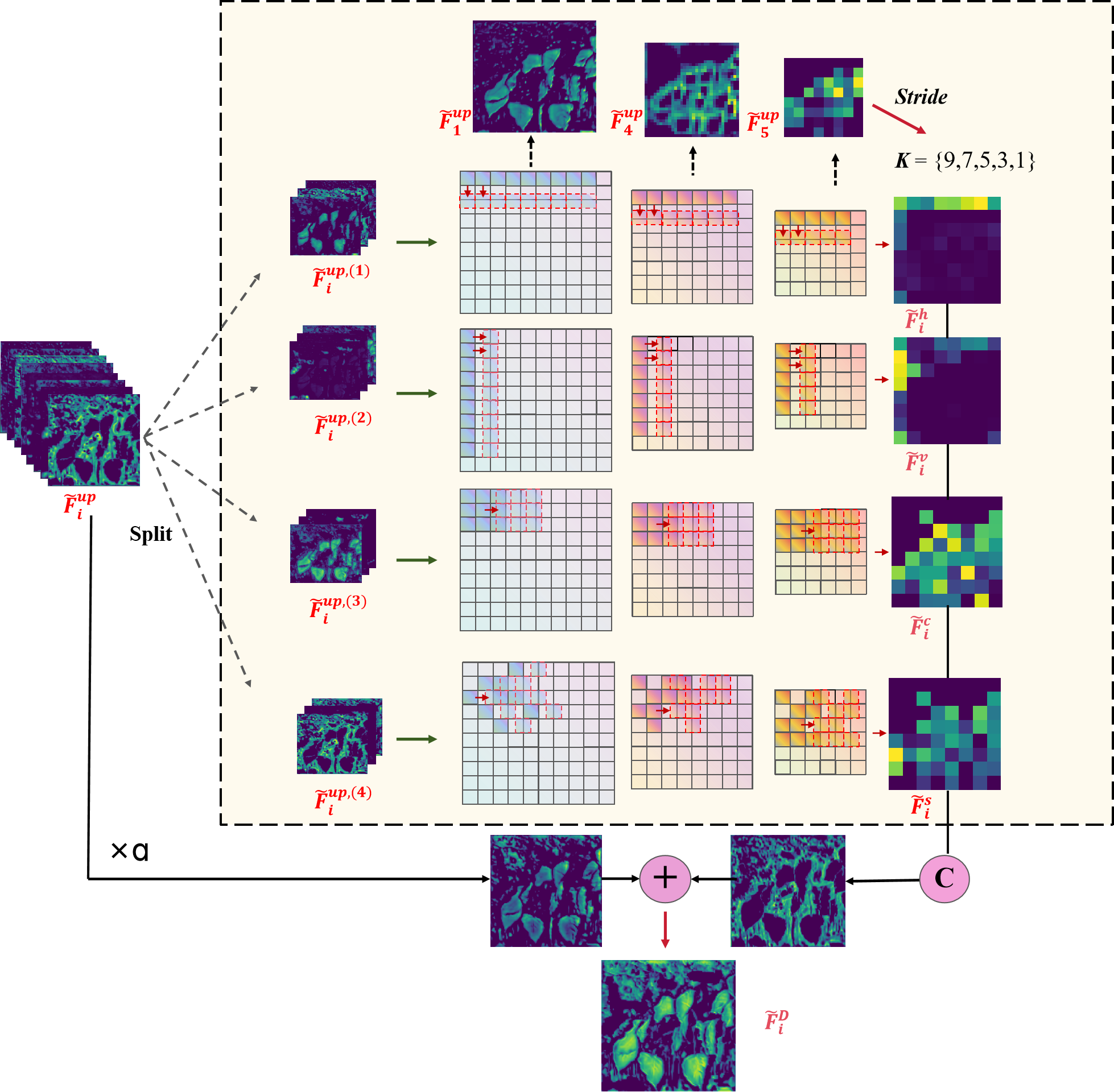}
	\caption{The feature visualization maps of dynamic asymmetric spatial perception.}
	\label{fig:6}
\end{figure*}

\subsection{Model Neck}
Multi-scale features play a crucial role in encoding objects with scale variations in object detection tasks. A common strategy for multi-scale feature extraction is to adopt the classic top-down and bottom-up Feature Pyramid Networks (FPN). However, these methods suffer from the loss or degradation of feature information, which weakens the fusion effect of non-adjacent levels. Inspired by Yang et al.\cite{70AFPN}, we leverage spatial features at different levels for dynamic interaction, forming the ASFF operation. Specifically, we first unify the resolutions through the identity alignment method between different levels. Downsampling is generally achieved through max - pooling and convolution, while upsampling is typically realized via linear interpolation. This process can be seen in Figure \ref{fig:4}. Then, by training a set of learnable weight parameters, we dynamically adjust the contribution of each feature layer at each level. The formula is simplified as follows:
\begin{align}
	&& X^{n \rightarrow i} = \mathrm{map}\!\left(X^{n}\right), &\\
	&& \tilde{F}_i
	= \alpha^{i} \cdot X^{n \rightarrow i}
	+ \cdots
	+ \gamma^{i} \cdot X^{i \rightarrow i}, &
\end{align}
where $\alpha^{i}, \ldots, \gamma^{i}$ are learnable weighting parameters, 
$\mathrm{map}(\cdot)$ denotes the identity alignment operation between different feature levels, and $\tilde{F}_i$ represents the feature map at the $i$-th level produced by the AFPN neck.\par

In addition, traditional mask branches typically employ a top-down, pixel-wise additive feature fusion strategy during the segmentation stage. However, this fusion operation is essentially identical to that adopted in the upstream Neck stage, which results in repeated weighting of low-level features throughout the feature fusion process. Such redundancy not only leads to feature duplication but also suppresses the effective utilization of high-resolution features, thereby hindering subsequent feature decoding. To address this issue, we further propose a Top-down Concatenation-decoder Feature Fusion (TCFU) strategy to mitigate feature redundancy and enhance high-resolution feature representation. 
\begin{figure}[t]
	\centering
	\includegraphics[width=1\textwidth]{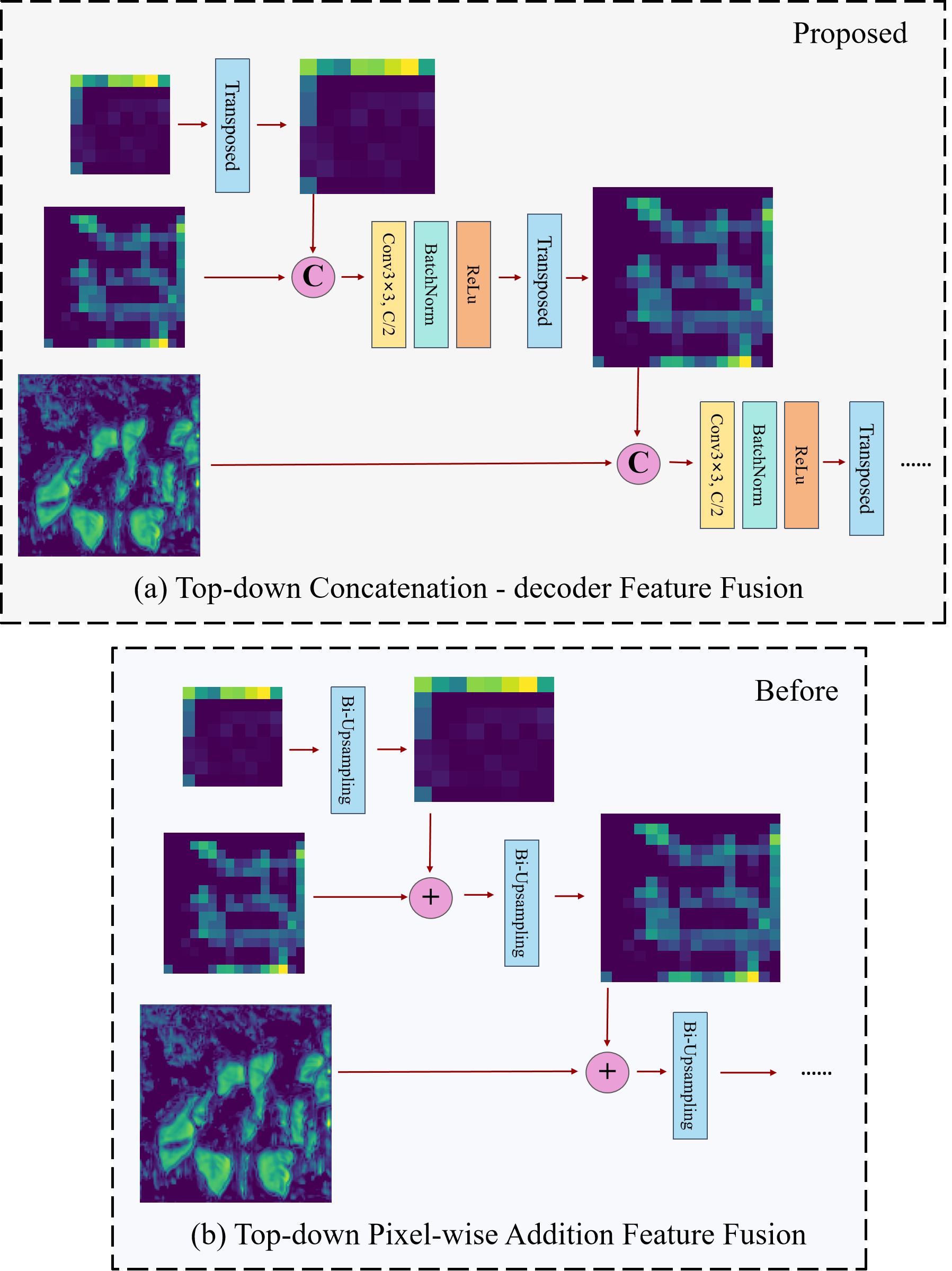}
	\caption{Different feature fusion strategies. (a) Top-down Concatenation - decoder Feature Fusion (b) Top-down Pixel-wise Addition Feature Fusion.}
	\label{fig:7}
\end{figure}
\subsection{Dynamic Anomalous Regression Head}
Dynamic Anomalous Regression Head (DARH) is a novel unified anchor-free detection head proposed by us, which can be used to replace the detection head for different models. The purpose of this head is to improve the detection end accuracy at a lightweight cost. A large number of experiments demonstrate that relying solely on convolutional branches with local dependencies is insufficient for effective feature integration at the neck stage, because leaf instances exhibit high variability in both shape and scale in natural environments. Since the leaf sway under the influence of wind, how to extract these deformation features poses a challenge. Additionally, high-density instance scenarios such as forestry leaves segmentation, present a difficulty for some transform-based decoders because their lack of “local view” (the downstream comparison experiments demonstrated this point). In contrast, DARH adopts a shared detection head across multiple feature scales with unified parameters, enabling consistent feature fusion while maintaining local sensitivity to deformation and scale variations.\par

The input to DARH is a feature map from the neck, denoted as $\tilde{F}_i \in \mathbb{R}^{C \times H \times W}$, where $\tilde{F}_i$ represents the feature at the $i$-th level of the feature pyramid. This feature map is first processed by a depthwise separable convolution to expand the channel dimension. The channel expansion is designed to enable downstream branches to adaptively and evenly select informative components from the feature maps in a learnable manner. The formulation is given as follows:

\begin{align}
	&& \tilde{F}_i^{up}
	= \partial\!\left(
	\mathrm{BN}\!\left(
	\mathrm{DWConv}\!\left(\tilde{F}_i\right)
	\right)
	\right), &
\end{align}
where $\partial(\cdot)$ denotes the ReLU activation function, $\mathrm{BN}(\cdot)$ denotes batch normalization, and $\mathrm{DWConv}(\cdot)$ represents the depthwise separable convolution operation.\par

Then, this extended feature is divided into four equal parts along the channel dimension through a linear mapping, ensuring that the input of each branch is the same as the original input. Then, we use the DASP branches to integrate different shapes of features and concatenate them to recover the channels’ depth compared to ${\tilde{F}_{i}^{up}}$, the details will be elaborate in next section. The overview formula is as follow:
\begin{align}
	&& \tilde{F}_i^{D}
	= \mathrm{Concat}\!\left(
	\mathrm{DASP}\!\left(
	\mathrm{Split}\!\left(\tilde{F}_i^{up}\right)
	\right)
	\right), &
\end{align}
where $\mathrm{Split}(\cdot)$ denotes a linear mapping that evenly partitions the input feature map along the channel dimension, $\mathrm{DASP}(\cdot)$ represents the set of shape-invariant convolution branches, and $\mathrm{Concat}(\cdot)$ concatenates the resulting branch features along the channel dimension. $\tilde{F}_i^{D}$ denotes the output feature map of the DASP module.\par
To preserve the long-range dependencies of features and eliminate the local field of view effects of convolution, we present dual-residual structure to restore features. The formula is as follows:
\begin{align}
	&& \tilde{F}_i^{out}
	= \beta \cdot \tilde{F}_i
	+ \partial\!\left(
	\mathrm{BN}\!\left(
	\Delta\!\left(
	\tilde{F}_i^{up} + \alpha \cdot \tilde{F}_i^{T}
	\right)
	\right)
	\right), &
\end{align}
where $\alpha$ and $\beta$ are dynamic weighting factors that control the contribution of feature restoration from different branches. $\Delta(\cdot)$ denotes a dimensionality expansion convolution used to align channel dimensions, $\mathrm{BN}(\cdot)$ denotes batch normalization, and $\partial(\cdot)$ denotes the ReLU activation function. Here, $\tilde{F}_i^{T}$ represents the concatenated feature map obtained from the DASP branches.
\par
Subsequently, the output feature map $\tilde{F}_i^{out}$ is used to predict bounding box coordinates and pixel-wise centerness scores. This regression process operates in parallel with a classification branch implemented by a standard convolutional decoder. Centerness is a scalar metric originally proposed by Tian et al.\cite{57con}, which serves as a proxy for estimating how close a pixel location is to the geometric center of its corresponding ground-truth bounding box. By assigning higher scores to pixels near the object center and suppressing those farther away, centerness guides the network to focus on more reliable candidate regions for object localization. This mechanism is widely adopted in anchor-free object detection frameworks. By jointly leveraging centerness prediction and bounding box regression, anchor-free detectors avoid the need for generating multi-scale anchor boxes at each pixel location and subsequently filtering them via NMS, thereby significantly improving computational efficiency. The formulation is given as follows:

\begin{align}
	&& centerness=  \sqrt{\frac{min({l}^{*},{r}^{*})}{max({l}^{*},{r}^{*})}\cdot\frac{min({t}^{*},{b}^{*})}{max({t}^{*},{b}^{*})}},&
\end{align}
where ${l}^{*}, {r}^{*}, {t}^{*}, {b}^{*}$ respectively refer to the distance between the pixel and the left, right, upper, and lower boundaries of the GT Bounding Box. $centerness$ is used to describe the degree of combing between pixel points and the center of the target.

Following the method proposed by them, a conditional controller is employed to dynamically predict the parameters of the mask head FCN. Specifically, for each detected instance, the controller predicts a total of 169 parameters based on the output features of the detection head, enabling instance-specific mask generation with a favorable trade-off between accuracy and computational cost. 
Motivated by this design, and leveraging the decoupled structure between the regression and classification branches in DARH, we likewise adopt the output feature map $\tilde{F}_i^{out}$ as controller features to dynamically predict mask weights. The formula is as follows:
\begin{align}
	&& \boldsymbol{\theta}_i 
	= \mathrm{Controller}\!\left(\tilde{F}_i^{out}\right), 
	\quad \boldsymbol{\theta}_i = [w_{i,1}, w_{i,2}, \ldots, w_{i,169}], &\\
	&& \left\{\mathbf{W}_i^{(1)}, \mathbf{W}_i^{(2)}, \mathbf{W}_i^{(3)}\right\}
	= \mathrm{Split}\!\left(\boldsymbol{\theta}_i\right), &\\
	&& \mathrm{Conv}_j(\cdot)
	= \mathrm{Conv}\!\left(\cdot;\mathbf{W}_i^{(j)}\right),
	\quad j \in \{1,2,3\}. &
\end{align}
where $\mathrm{Controller}(\cdot)$ denotes the conditional controller module that maps the detection feature $\tilde{F}_i^{out}$ to an instance-specific parameter vector $\boldsymbol{\theta}_i \in \mathbb{R}^{169}$. 
The vector $\boldsymbol{\theta}_i = [w_{i,1}, w_{i,2}, \ldots, w_{i,169}]$ contains the dynamically predicted convolutional parameters for the $i$-th instance. 
The operator $\mathrm{Split}(\cdot)$ partitions $\boldsymbol{\theta}_i$ into three parameter groups $\{\mathbf{W}_i^{(1)}, \mathbf{W}_i^{(2)}, \mathbf{W}_i^{(3)}\}$, which are used to parameterize the dynamic convolution layers in the mask head. 
Here, $\mathrm{Conv}_j(\cdot)$ denotes the $j$-th dynamic convolution operation with instance-specific weights $\mathbf{W}_i^{(j)}$, where $j \in \{1,2,3\}$.

\subsection{Dynamic Asymmetric Spatial Perception}
Dynamic Asymmetric Spatial Perception is specifically designed to capture features of different scales and shapes. This module can be integrated with any other downstream tasks, including but not limited to leaf instance segmentation. The structual diagram can be seen in Figure \ref{fig:5}. Next, we will demonstrate the function of this module through feature map visualization. \par
For forestry scenarios, the leaves exhibit irregular swaying. This can result in horizontal, vertical, and concentrated distribution of features. To capture the features of these anomalies, we designed a shape invariant convolution module. Firstly, feature maps from different levels are extracted through horizontal convolution, vertical convolution, deep convolution, and deformable convolution. The specific process can be seen in Figure \ref{fig:6}. Let $\tilde{F}_i^{up}$ denote the upsampled feature map at pyramid level $i$. We split $\tilde{F}_i^{up}$ into multiple feature groups and feed them into the shape-invariant convolution branches. The formulas are as follows:
\begin{align}
	&& \left\{\tilde{F}_i^{up,(1)},\tilde{F}_i^{up,(2)},\tilde{F}_i^{up,(3)},\tilde{F}_i^{up,(4)}\right\}
	= \mathrm{Split}\!\left(\tilde{F}_i^{up}\right), &\\
	&& \tilde{F}_i^{h} = \mathrm{HorizontalConv}\!\left(\tilde{F}_i^{up,(1)}\right), &\\
	&& \tilde{F}_i^{v} = \mathrm{VerticalConv}\!\left(\tilde{F}_i^{up,(2)}\right), &\\
	&& \tilde{F}_i^{c} = \mathrm{DeepConv}\!\left(\tilde{F}_i^{up,(3)}\right), &\\
	&& \tilde{F}_i^{s} = \mathrm{DeformConv}\!\left(\tilde{F}_i^{up,(4)}\right). &
\end{align}
where $\mathrm{HorizontalConv}(\cdot)$ and $\mathrm{VerticalConv}(\cdot)$ denote the horizontal and vertical convolution branches, respectively, $\mathrm{DeepConv}(\cdot)$ denotes the deep convolution branch, and $\mathrm{DeformConv}(\cdot)$ denotes the deformable convolution branch. The resulting feature maps $\tilde{F}_i^{h}$, $\tilde{F}_i^{v}$, $\tilde{F}_i^{c}$, and $\tilde{F}_i^{s}$ correspond to the outputs of the horizontal, vertical, deep, and deformable branches, respectively.\par

The kernel size of horizontal and vertical convolutions is affected by the resolution downsampling ratio: $Stride$. For models with high downsampling rates, we will use a lower convolution kernel size $K$. In this experiment, we let $K$ be equally to 3. Then, concatenate the multivariate feature maps and apply a linear projection to obtain the preliminary fusion feature layer: ${{\tilde{F}}}_{i}^{T}$. The formula is as follows:
\begin{align}
	&& \tilde{F}_i^{T}
	= \mathrm{Linear}\!\left(
	\mathrm{Concat}\!\left(
	\tilde{F}_i^{h}, \tilde{F}_i^{v}, \tilde{F}_i^{c}, \tilde{F}_i^{s}
	\right)
	\right), &
\end{align}
where $\mathrm{Concat}(\cdot)$ concatenates the branch feature maps $\{\tilde{F}_i^{h}, \tilde{F}_i^{v}, \tilde{F}_i^{c}, \tilde{F}_i^{s}\}$ along the channel dimension, and $\mathrm{Linear}(\cdot)$ denotes a linear projection that maps the concatenated features to a fixed channel dimension. By default, the output channel dimension is set to $256 \times 4$, thereby restoring the feature depth after channel partitioning.
\par

Finally, we design a residual structure to preserve the original information of the feature map. Specifically, the upsampled feature map $\tilde{F}_i^{up}$ is scaled by a fixed factor $\alpha = 0.3$ and added to the preliminary fused feature representation, forming the final output.

\begin{align}
	&&{{\tilde{F}}}_{i}^{D} = {{\tilde{F}}}_{i}^{up} \cdot \alpha + {{\tilde{F}}}_{i}^{T},&
\end{align}
where ${{\tilde{F}}}_{i}^{D}$ represents the final output feature map of the DSAP module. The feature layer is then input into the second residual branch to resotre original information for futher, and finally the feature will be input into TCFU model for feature fuision to enhance the mask end precision.

\subsection{Top-down Concatenation - decoder Feature Fusion}
This subsection focuses on exploring the impacts of two different feature fusion strategies on feature integration. Figure \ref{fig:7} shows schematic diagrams of the two feature strategies. (b) illustrates Top-down Pixel-wise Addition Feature Fusion, a commonly used operation for feature fusion after the Neck stage. Its core concept involves upsampling low-level features and integrating them with high-level features through pixel-wise addition to capture long-range dependencies. The formula is as follows:
\begin{align}
	&& F_{i+1}' = \mathcal{U}\!\left(F_i'\right) + F_{i+1}, &
\end{align}
where $\mathcal{U}(\cdot)$ denotes the bilinear interpolation operation for feature upsampling, $F_i$ represents the feature map at the $i$-th pyramid level, $F_i'$ denotes the fused feature map propagated from the previous level, and $F_{i+1}$ denotes the original feature map at level $i+1$. Accordingly, $F_{i+1}'$ represents the fused feature map at level $i+1$. \par

The original intention behind this approach is that low-level and high-level features focus on different aspects: low-level features contain more edge and texture details, while high-level features provide more holistic contour descriptions. Adding them together seems to address the issue of insufficient edge details in high-level features. However, upon re-examining this scheme, it becomes evident that it leads to feature redundancy. Specifically, the texture details of low-level features are repeatedly weighted at the Neck stage. If top-down pixel-wise addition continues at the Mask stage, the weights of these details in the response feature layer become excessively high, causing the model to overlook holistic contour features.\par
As illustrated in Fig.~\ref{fig:7}(a), we introduce the Top-down Concatenation-decoder Feature Fusion (TCFU) module, which is designed to rebalance the contributions of hierarchical features. 
This module adopts a recursive fusion strategy. Specifically, for each layer except the first, the input feature ${F}_{i}^{\prime}$ originating from the concatenation of the previous layer is first processed by a convolution with $C/2$ filters to reduce its channel dimensionality. Subsequently, a series of transposed convolutions are applied to upsample the feature map to match the spatial resolution of the next layer. Finally, the upsampled features are concatenated with the corresponding features of the next layer and recursively fed into the subsequent fusion stage. The formula is as follows:
\begin{align}
	&& F_{i+1}' 
	= \mathrm{Concat}\!\left(
	\mathcal{T}\!\left(
	\Delta\!\left(F_i'\right)
	\right),
	F_{i+1}
	\right), &
\end{align}
where $\Delta(\cdot)$ denotes a convolutional projection used for channel dimensionality reduction, $\mathcal{T}(\cdot)$ denotes the transposed convolution operation for spatial upsampling, and $\mathrm{Concat}(\cdot,\cdot)$ represents channel-wise feature concatenation. Here, $F_i'$ denotes the fused feature map propagated from the previous pyramid level, and $F_{i+1}'$ represents the resulting fused feature map at level $i+1$.
\par
The advantage of this approach is that it assigns trainable convolutional parameters to the feature weighting process, rather than simply emphasizing through straightforward addition. Direct addition implies a human prior that "details are more important," which is unfair in certain scenarios. First, the model integrates critical information from the channel dimensions of the previous layer through dimensionality reduction. It then emphasizes the ROI in spatial features via transposed convolutions. Finally, concatenating with the next layer and recursively helps reduce feature redundancy.

\subsection{Loss Function}
\label{sec:loss}

The proposed framework jointly predicts object class probabilities, bounding box coordinates, centerness scores, and instance segmentation masks. The overall training objective is formulated as a multi-task loss:
\begin{align}
	\mathcal{L}
	&= \lambda_{\mathrm{cls}} \mathcal{L}_{\mathrm{cls}}
	+ \lambda_{\mathrm{bbox}} \mathcal{L}_{\mathrm{bbox}}
	+ \lambda_{\mathrm{cent}} \mathcal{L}_{\mathrm{cent}}
	+ \lambda_{\mathrm{mask}} \mathcal{L}_{\mathrm{mask}},
\end{align}
where the weights for the classification and centerness losses are set to 0.5, and those for the bounding box regression and mask losses are set to 2.0.

\begin{figure*}[t]
	\centering
	\includegraphics[width=0.8\textwidth]{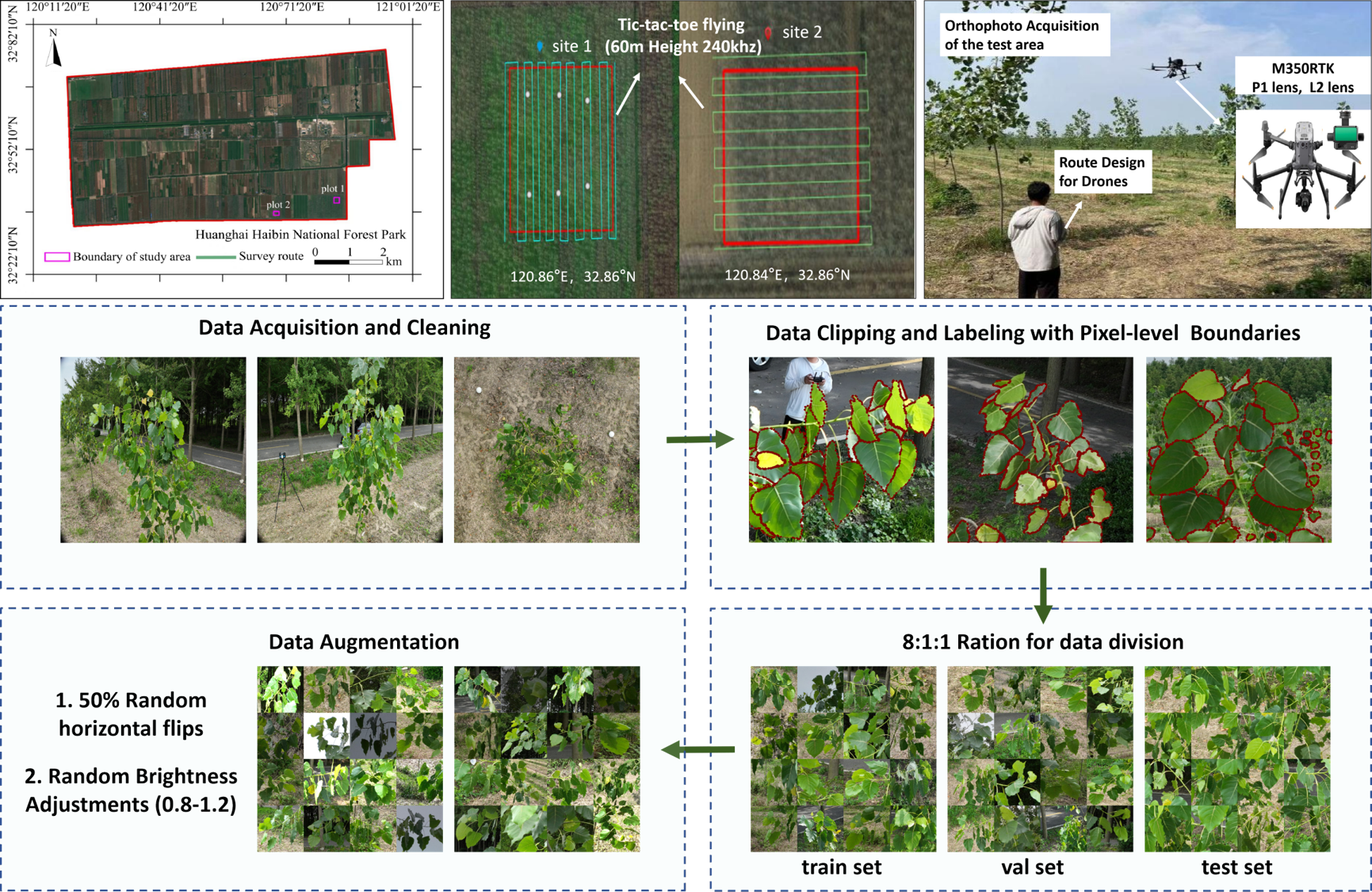}
	\caption{Study area and dataset production.}
	\label{fig:8}
\end{figure*}
\begin{figure*}[t]
	\centering
	\includegraphics[width=0.6\textwidth]{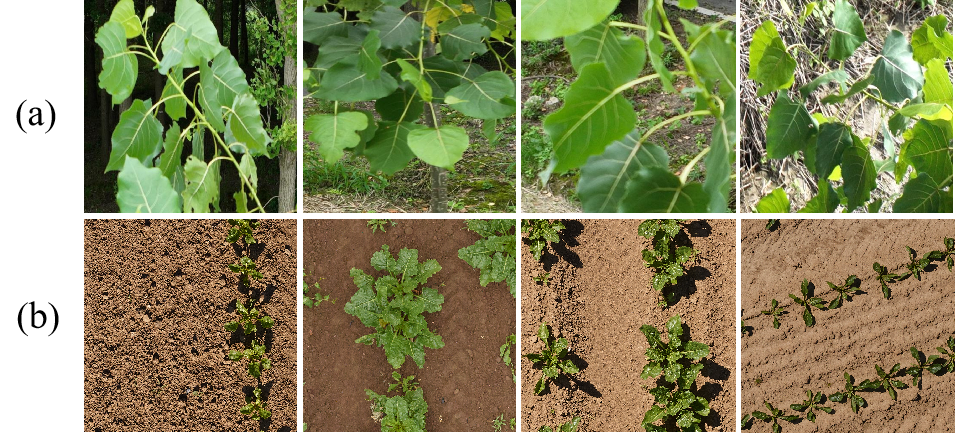}
	\caption{Examples of different scenarios. (a) Forestry Dataset: Poplar-leaf, (b) Agriculture Dataset: PhenoBench.}
	\label{fig:9}
\end{figure*}
\paragraph{Classification loss.}
To address the severe foreground--background imbalance in dense instance segmentation especially in forestry scene, we adopt the focal loss:
\begin{align}
	\mathcal{L}_{\mathrm{cls}}
	&= - \frac{1}{N}
	\sum_{i=1}^{N}
	\alpha \left(1 - p_i\right)^{\gamma} \log(p_i),
\end{align}
where $p_i$ denotes the predicted probability for the leaf at location $i$,
$\alpha$ is a class-balancing factor, $\gamma$ is the focusing parameter, and $N$ is the
number of training samples. In our experiments, $\alpha=0.25$ and $\gamma=2.0$.

\paragraph{Bounding box regression loss.}
Bounding box localization is supervised using the Generalized Intersection-over-Union
(GIoU) loss. Given a predicted bounding box $\hat{\mathbf{b}}$ and its corresponding
ground-truth box $\mathbf{b}$, the GIoU is defined as:
\begin{align}
	\mathrm{GIoU}(\mathbf{b}, \hat{\mathbf{b}})
	&= \mathrm{IoU}(\mathbf{b}, \hat{\mathbf{b}})
	- \frac{\left| C \setminus (\mathbf{b} \cup \hat{\mathbf{b}}) \right|}{|C|},
\end{align}
where $C$ denotes the smallest enclosing box covering both $\mathbf{b}$ and
$\hat{\mathbf{b}}$, and $|\cdot|$ represents the area of a region.
The bounding box regression loss is then formulated as:
\begin{align}
	\mathcal{L}_{\mathrm{bbox}}
	&= 1 - \mathrm{GIoU}(\mathbf{b}, \hat{\mathbf{b}}).
\end{align}

\paragraph{Centerness loss.}
Following anchor-free detection frameworks, a centerness branch is introduced to suppress
low-quality predictions far from leaves' centers. The ground-truth centerness target is
defined as before in Eq(6). 
\begin{align}
	\mathcal{L}_{\mathrm{cent}}
	&= - \frac{1}{N}
	\sum_{i=1}^{N}
	\left[
	c_i \log(\hat{c}_i)
	+ (1 - c_i)\log(1 - \hat{c}_i)
	\right].
\end{align}
where $c$ denotes centerness here, $\hat{c}$ denotes the prediction value.

\paragraph{Mask loss.}
For instance segmentation, we employ the Dice loss to directly optimize the overlap
between predicted and ground-truth masks:
\begin{align}
	\mathcal{L}_{\mathrm{mask}}
	&= 1 -
	\frac{
		2 \sum_{j} \hat{m}_j m_j + \epsilon
	}{
		\sum_{j} \hat{m}_j^2 + \sum_{j} m_j^2 + \epsilon
	},
\end{align}
where $\hat{m}_j$ and $m_j$ denote the predicted and ground-truth mask values at pixel $j$,
and $\epsilon$ is a small constant for numerical stability.

\begin{figure*}[t]
	\centering
	\includegraphics[width=1\textwidth]{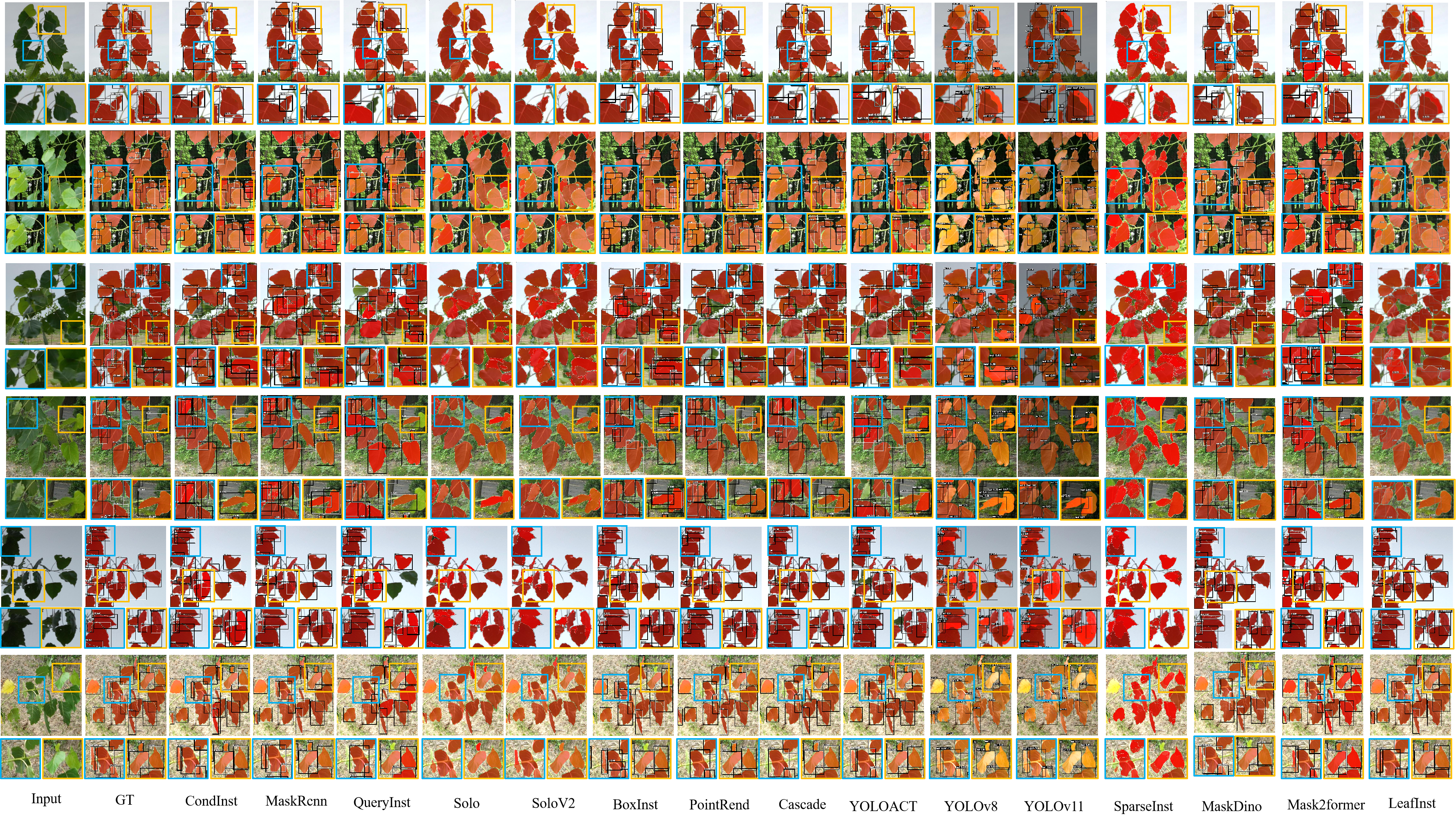}
	\caption{Qualitative comparison results with different models on Poplar-leaf}
	\label{fig:10}
\end{figure*}
\begin{figure*}[t]
	\centering
	\includegraphics[width=1\textwidth]{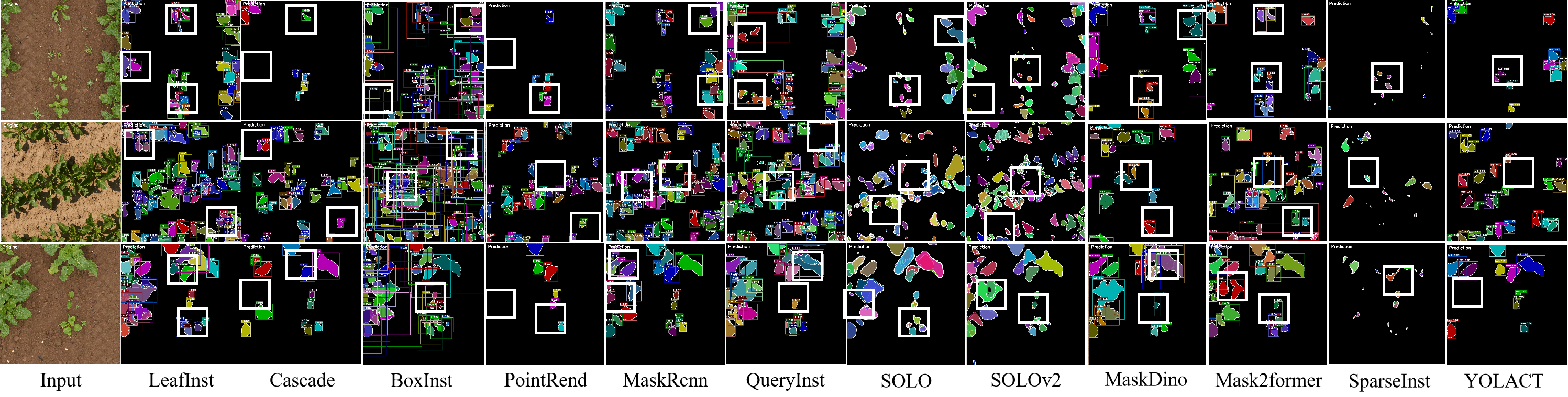}
	\caption{Qualitative comparison results with different models on Phenobench.}
	\label{fig:11}
\end{figure*}
\section{Experiment Settings}
This section provides a complete introduction to our study area, dataset production process, evaluation metrics, and experimental details.

\subsection{Dataset and Study Area}
\textbf{Poplar-leaf}: Acquiring High-Quality Datasets to Address Forestry Breeding Challenges. To tackle the challenges in forestry breeding, we selected the poplar plantation in the Huanghai Seaside National Forest Park, Dongtai City, Jiangsu Province, China, located at $32^{\circ}51'N$, $120^{\circ}50'E$, as our research area. The specific information is shown in Figure \ref{fig:8}. Spanning over 13,000 km$^2$, this region features dense poplar forests with luxuriant branches. Composed of 3-year-old poplar seedlings with a breast diameter of 3-8 cm and relatively sparse branches, the plantation includes two sub-areas with distinct environmental conditions: soil and water sources, providing a basis for comparative data analysis under different scenarios. Data collection was conducted using a DJI M350RTK drone equipped with a P1 camera. The drone flew at a 5-meter altitude along pre-set survey routes and captured multi-angle images around individual trees to obtain high-resolution aerial imagery from diverse perspectives. For annotation, we used LabelMe software to perform pixel-level instance labeling on each tree, covering 1,202 leaf branches and 19,876 leaf instances, with mask and bounding box information for each leaf. An 8-person team completed the labeling task over 7 days, using the built-in SAM model for initial processing followed by fine-grained manual verification and correction. To ensure coverage of all forest areas, we selected images with a repetition rate $\leq$30\% for labeling. The dataset consists of 1,202 labeled images, which are split into an 8:1:1 ratio for training, validation, and testing, each with a resolution of 1,024 $\times$ 1,024 pixels, and guarantees that 80\% of images contain $\geq$8 leaf instances per frame. Meanwhile, we implemented data augmentation using two strategies: 1. Random horizontal flipping with a probability of 50\%, and 2. Random illumination adjustment within the range of 0.8 to 1.1. Additionally, we provide 40.8 GB of unlabeled leaf images covering nearly all young poplar trees in the region, suitable for transfer learning and unsupervised network training. As the first large-scale dataset for leaf instance segmentation in forestry, Poplar-leaf establishes a foundational resource for the field. Unlike agricultural crop leaf datasets, which focus on relatively single species, this dataset captures leaf instances under natural light and growth conditions from diverse drone angles. It offers forestry researchers abundant samples to study leaf growth patterns, morphological characteristics, and physiological ecology. Through analysis of numerous leaf instances, insights into tree growth mechanisms can be gained, providing a robust foundation for forestry breeding, cultivation, and related research. \par

\textbf{PhenoBench}: A large-scale public dataset specifically designed for semantic image interpretation in the agricultural field, proposed by Weyler et al\cite{72pb}. The dataset provides RGB images recorded by drones equipped with high-resolution cameras under real field conditions. They used a DJI M600 with a Phase One iXM - 100 camera, which was equipped with an 80 mm RSM fixed - focus lens and mounted on a gimbal to obtain motion - stabilized RGB images with a resolution of 11664  $\times$ 8750 pixels. The drone flew at an altitude of approximately 21 m, and the ground sampling distance (GSD) was 1 mm/pixel. After manual cropping and annotation by the author's team, each image patch was sized at 1024 $\times$ 1024 pixels. They provided leaf segmentation instances of sugar beet crops and weeds. We selected the leaf instance part to verify the model's transferability.\par

Sample images from the two datasets are shown in Figure \ref{fig:9}. Compared with the public dataset, our dataset covers images captured from diverse UAV viewing angles, enabling the observation of open-scene forestry images with varying leaf deformations and perspectives. In contrast, PhenoBench only provides orthorectified remote sensing imagery from a single viewing angle. 
\begin{table*}[t]
	\centering
	\caption{Poplar-Leaf Validation Results (\%): 
		\colorbox{redbg}{\phantom{x}\phantom{x}}: 1st place, 
		\colorbox{yellowbg}{\phantom{x}\phantom{x}}: 2nd place, 
		\colorbox{bluebg}{\phantom{x}\phantom{x}}: 3rd place.}
	\label{tab:val_results}
	\begin{tabularx}{\textwidth}{X *{6}{C}}
		\toprule
		Methods & seg/mAP & seg/AP50 & seg/AP75 & box/mAP & box/AP50 & box/AP75 \\
		\midrule
		Queryinst & 56.9 & 76.4 & 62.1 & 55.3 & 75.3 & 60.2 \\
		CascadeRcnn & 63.6 & 84.7 & \colorbox{yellowbg}{72.3} & 64.1 & 84.6 & \colorbox{yellowbg}{71.4} \\
		MaskRcnn & 61.8 & 86.5 & 68.5 & 61.5 & 86.4 & 67.5 \\
		Solo & 56.8 & 80.9 & 63.0 & -- & -- & -- \\
		SoloV2 & \colorbox{bluebg}{65.3} & 86.7 & \colorbox{bluebg}{71.6} & -- & -- & -- \\
		PointRend & 64.4 & 86.5 & 70.5 & 62.2 & 86.4 & \colorbox{bluebg}{69.4} \\
		BoxInst & 58.2 & 85.3 & 64.1 & 62.3 & 85.8 & 68.6 \\
		YOLOV8 & 61.9 & \colorbox{bluebg}{87.6} & -- & \colorbox{bluebg}{64.5} & \colorbox{yellowbg}{87.8} & -- \\
		YOLOV11 & 62.2 & \colorbox{yellowbg}{88.1} & -- & \colorbox{yellowbg}{64.6} & \colorbox{bluebg}{87.6} & -- \\
		YOLACT & 54.0 & 77.1 & 57.8 & 49.0 & 76.6 & 55.1 \\
		Mask2former & 56.4 & 75.6 & 60.9 & 53.7 & 72.4 & 56.9 \\
		MaskDino & \colorbox{yellowbg}{65.3} & 84.9 & 70.9 & 61.1 & 82.2 & 64.7 \\
		SparseInst & 43.1 & 66.5 & 45.6 & -- & -- & -- \\
		LeafInst & \colorbox{redbg}{68.4} & \colorbox{redbg}{88.9} & \colorbox{redbg}{73.8} & \colorbox{redbg}{65.6} & \colorbox{redbg}{87.8} & \colorbox{redbg}{72.5} \\
		\bottomrule
	\end{tabularx}
\end{table*}

\begin{table*}[t]
	\centering
	\caption{Poplar-Leaf Test Results (\%): 
		\colorbox{redbg}{\phantom{x}\phantom{x}}: 1st place, 
		\colorbox{yellowbg}{\phantom{x}\phantom{x}}: 2nd place, 
		\colorbox{bluebg}{\phantom{x}\phantom{x}}: 3rd place.}
	\label{tab:test_results}
	\begin{tabularx}{\textwidth}{X *{6}{C}}
		\toprule
		Methods & seg/mAP & seg/AP50 & seg/AP75 & box/mAP & box/AP50 & box/AP75 \\
		\midrule
		Queryinst & 58.2 & 79.7 & 64.4 & 55.7 & 79.2 & 59.5 \\
		CascadeRcnn & 62.3 & 83.0 & 69.3 & 62.4 & 83.0 & \colorbox{yellowbg}{69.8} \\
		MaskRcnn & 62.1 & \colorbox{yellowbg}{90.7} & 68.8 & 61.2 & 89.8 & 65.2 \\
		Solo & 57.9 & 84.3 & 65.1 & -- & -- & -- \\
		SoloV2 & \colorbox{yellowbg}{64.9} & 88.6 & \colorbox{yellowbg}{71.7} & -- & -- & -- \\
		PointRend & \colorbox{bluebg}{63.6} & 85.0 & \colorbox{bluebg}{70.5} & 61.7 & 84.6 & \colorbox{bluebg}{68.2} \\
		BoxInst & 57.9 & 87.9 & 64.9 & 61.2 & 88.1 & 66.4 \\
		YOLOV8 & 62.0 & \colorbox{bluebg}{89.7} & -- & \colorbox{yellowbg}{65.7} & \colorbox{redbg}{90.5} & -- \\
		YOLOV11 & 62.9 & 89.3 & -- & \colorbox{bluebg}{64.9} & \colorbox{yellowbg}{90.4} & -- \\
		YOLACT & 50.3 & 78.6 & 53.4 & 45.7 & 79.2 & 47.7 \\
		Mask2former & 58.0 & 78.6 & 53.4 & 55.1 & 75.6 & 58.0 \\
		MaskDino & 63.5 & 85.7 & 68.9 & 57.8 & 81.1 & 60.1 \\
		SparseInst & 43.5 & 68.9 & 45.5 & -- & -- & -- \\
		LeafInst & \colorbox{redbg}{70.0} & \colorbox{redbg}{91.9} & \colorbox{redbg}{77.0} & \colorbox{redbg}{65.8} & \colorbox{yellowbg}{90.4} & \colorbox{redbg}{71.0} \\
		\bottomrule
	\end{tabularx}
\end{table*}

\begin{table*}[t]
	\centering
	\caption{Quantitative Comparison of Different Methods on PhenoBench (\%): \colorbox{redbg}{\phantom{x}\phantom{x}}: 1st place, \colorbox{yellowbg}{\phantom{x}\phantom{x}}: 2nd place.}
	\label{tab:phenobench}
	\begin{tabularx}{\textwidth}{X *{4}{C}}
		\toprule
		Methods & box/mAP & box/AP50 & seg/mAP & seg/AP50 \\
		\midrule
		LeafInst & \colorbox{redbg}{52.7} & \colorbox{redbg}{82.3} & \colorbox{yellowbg}{50.2} & \colorbox{redbg}{80.5} \\
		YOLOv8 & 48.6 & 78.4 & 45.7 & 76.0 \\
		YOLOv11 & 47.5 & 77.7 & 45.6 & 75.9 \\
		MaskDINO & \colorbox{yellowbg}{49.3} & \colorbox{yellowbg}{80.2} & \colorbox{redbg}{51.0} & \colorbox{yellowbg}{80.3} \\
		\bottomrule
	\end{tabularx}
\end{table*}

\begin{figure*}[t]
	\centering
	\includegraphics[width=0.8\textwidth]{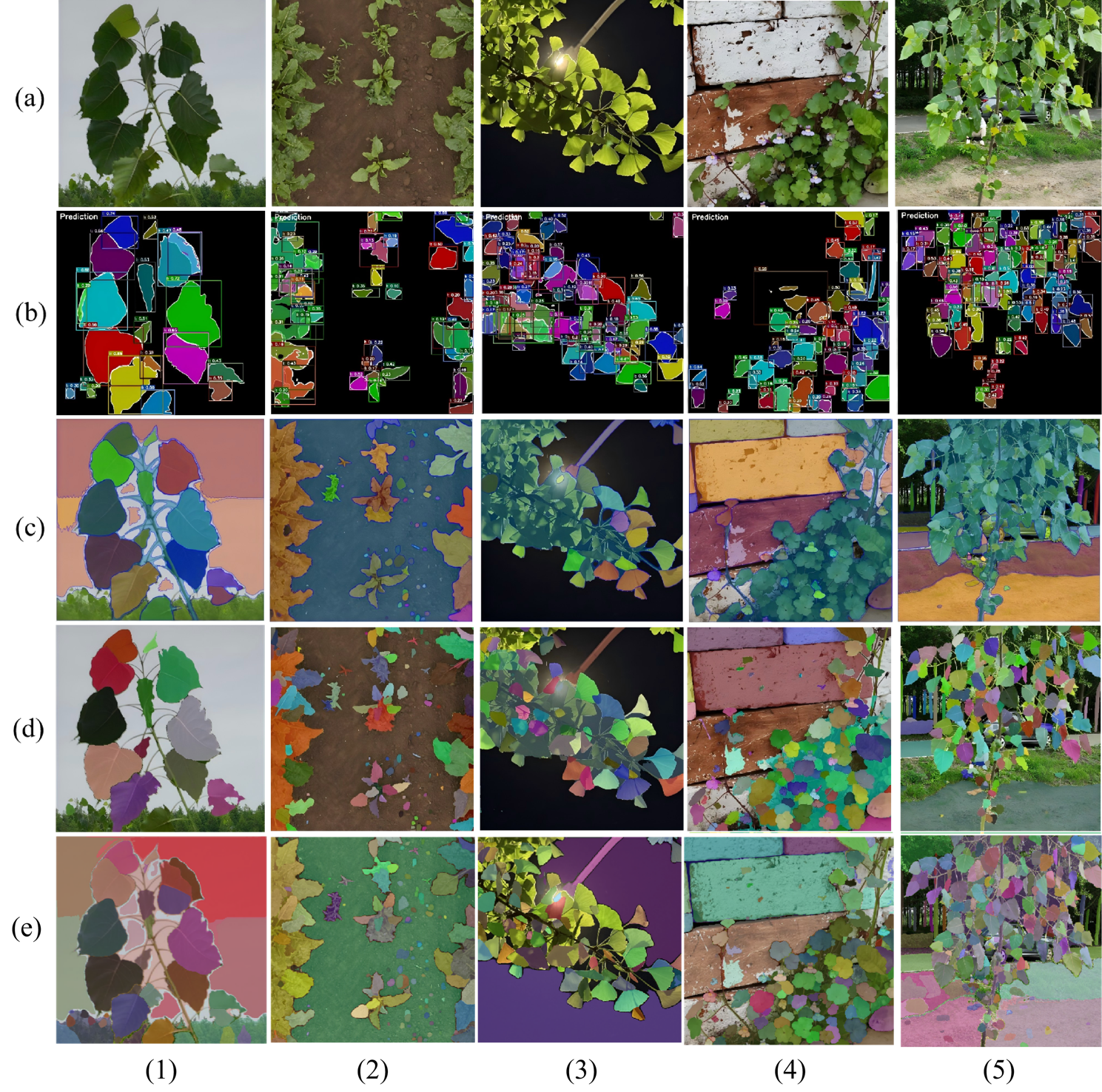}
	\caption{Visualization comparison with Visual Foundation Models. (a) Input (b) LeafInst (c) SAM (d) Fast-SAM (e) semantic-SAM}
	\label{fig:12}
\end{figure*}
\subsection{Evaluation Metrics}

In the quantitative comparison, we adopted standard COCO Metrics to measure the model's performance. Specifically, we applied Box and segmentation metrics to evaluate the model's performance, respectively. The Average Precision (AP) metrics under different Intersection of Union (IoU) thresholds were used to quantitatively analyze the model's accuracy. IoU measures the degree of overlap between two regions by calculating the ratio of the intersection area to the union area of the two regions. In object detection tasks, these two regions are typically the prediction box (the target position predicted by the model) and the ground-truth box (the actual target position). In image segmentation tasks, they are the predicted segmentation region and the ground-truth segmentation region. The formula is as follows:
\begin{align}
	&&IoU = \frac{Pred \bigcap GT}{Pred \bigcup GT},&
\end{align}
where $Pred$ denotes the predication results, $GT$ denotes the ground truth of the images. The formula is as follows: 
\begin{align}
	&&Precision = \frac{TP}{TP+FP},&\\
	&&Recall = \frac{TP}{TP+FN},&
\end{align}
where $TP$ (True Positive) represents the instances correctly identified and classified by the model. $FP$ (False Positive) refers to the instances incorrectly identified as positive by the model but are actually negative. $FN$ (False Negative) indicates the actual positive instances that the model fails to detect or classify correctly.\par

In the COCO evaluation criteria, the mean average precision (mAP) is obtained by averaging the AP values across multiple intersection-over-union (IoU) thresholds ranging from 0.50 to 0.95 with a step of 0.05, providing a comprehensive assessment of a model's object detection and segmentation performance. Additionally, AP50 and AP75 represent the average precision at fixed IoU thresholds of 0.50 and 0.75, respectively. The difference between these two metrics reflects the model's stability under varying precision requirements, aiding in evaluating its adaptability to different application scenarios. \par

In addition to the standard COCO metrics, we further introduce a manual visual inspection--based leaf-level accuracy to better evaluate model performance in complex forestry scenarios. In dense leaf canopies with severe occlusion and frequent overlaps, strict IoU-based criteria may underestimate perceptually correct predictions, especially when minor boundary deviations exist. Moreover, this manual visual inspection metric is also applicable for quantifying qualitative comparison results of SAM-based models, which do not provide instance-level outputs or IoU-based evaluation interfaces. In such cases, manual inspection becomes the only feasible way to assess whether individual leaf instances are correctly identified. In addition, this metric is well suited to broader zero-shot transfer scenarios, where precise instance-level contour annotations are unavailable. Let $\mathcal{G}=\{G_1,\dots,G_{N_G}\}$ denote the set of ground-truth leaf instances, and $\mathcal{P}=\{P_1,\dots,P_{N_P}\}$ denote the set of predicted leaf instances. A predicted instance is considered correctly identified if, according to manual visual inspection, it corresponds to a unique ground-truth leaf in terms of spatial location and leaf identity, under a one-to-one matching constraint. The total number of visually correct detections is denoted as $TP_{\mathrm{vis}}$. The manual visual inspection accuracy is defined as:
\begin{align}
	&& Accuracy_{\mathrm{vis}} = \frac{TP_{\mathrm{vis}}}{N_G},&
\end{align}
where $N_G$ denotes the total number of ground-truth leaf instances. 
This metric measures the proportion of leaves that are successfully recognized by the model from a human perceptual perspective.
By complementing IoU-based quantitative metrics with manual visual inspection, we provide a more comprehensive evaluation of model robustness and practical applicability, particularly for fine-grained leaf instance segmentation under varying scales, viewpoints, and illumination conditions.\par
\subsection{Experiments Setups}
All experiments were conducted on a system equipped with an AMD Ryzen 9 7950X 16-core processor and an NVIDIA GeForce RTX 3090 graphics card (24GB VRAM), running the Ubuntu 22.04 operating system. The experiments utilized CUDA 11.8 and PyTorch 2.1.0, with model training and inference implemented based on the mmdetection and ultralytics frameworks. All models were trained for 36 epochs or 8064 iterations (the two are equivalent). The SGD optimizer was used with an initial learning rate of 0.005, a batch size of 4, and linear learning rate warm-up with a start\_factor of 0.01 over 400 iterations. The learning rate decay exponent was 0.5, with decay applied at the 2/3 and 4/5 points of training. The input image resolution was 1024 $\times$ 1024. Slightly different training strategies were applied to different models to optimize their performance. The backbone used was ResNet50, loaded with ImageNet pretrained weights \cite{61res}. All metrics were calculated using built-in scripts from official libraries. The specific training strategies for our model have been open-sourced at Github.
\begin{figure*}[h]
	\centering
	\includegraphics[width=0.8\textwidth]{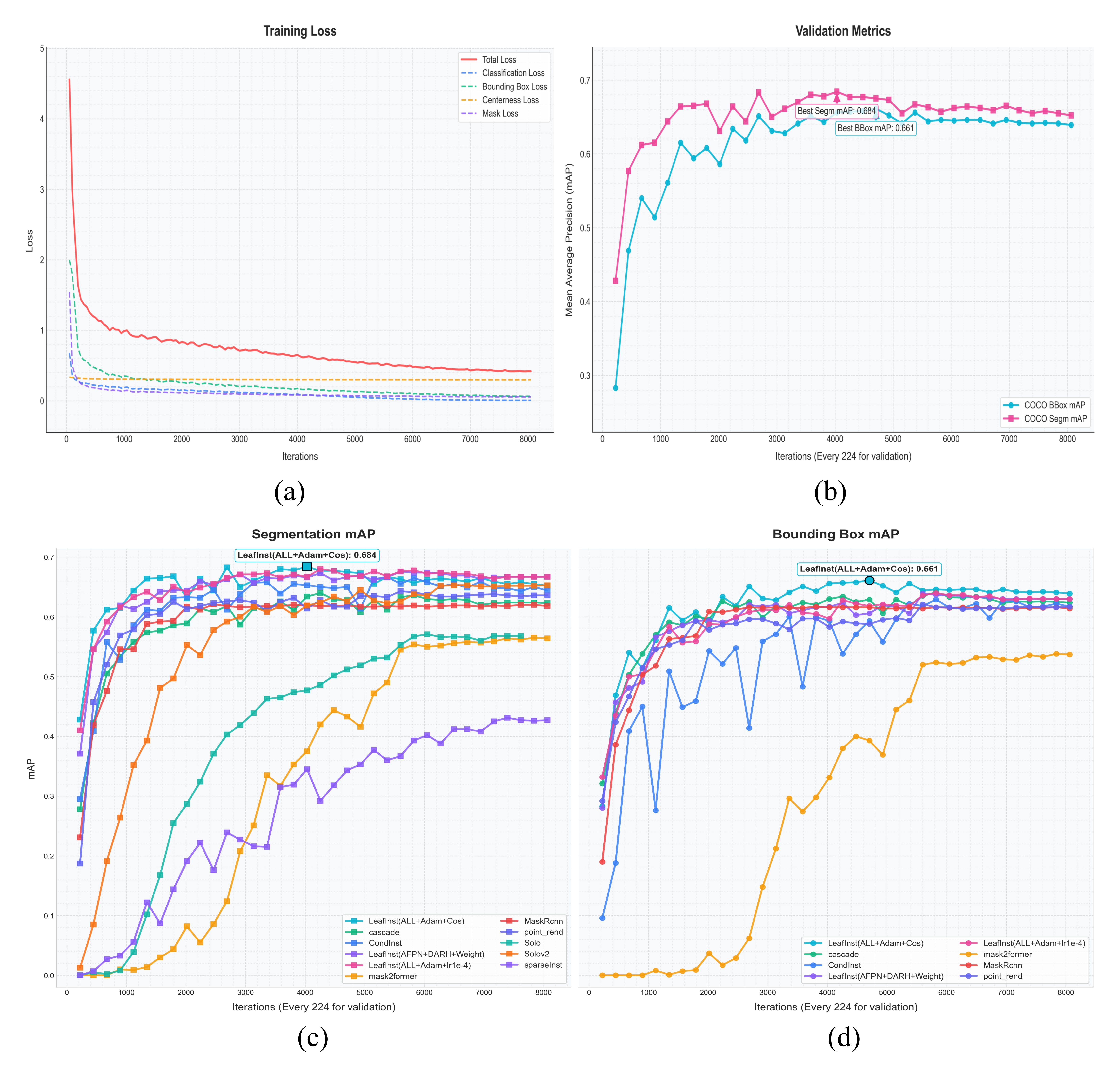}
	\caption{Accuracy and convergence analysis.}
	\label{fig:13}
\end{figure*}
\begin{figure*}[t]
	\centering
	\includegraphics[width=0.8\textwidth]{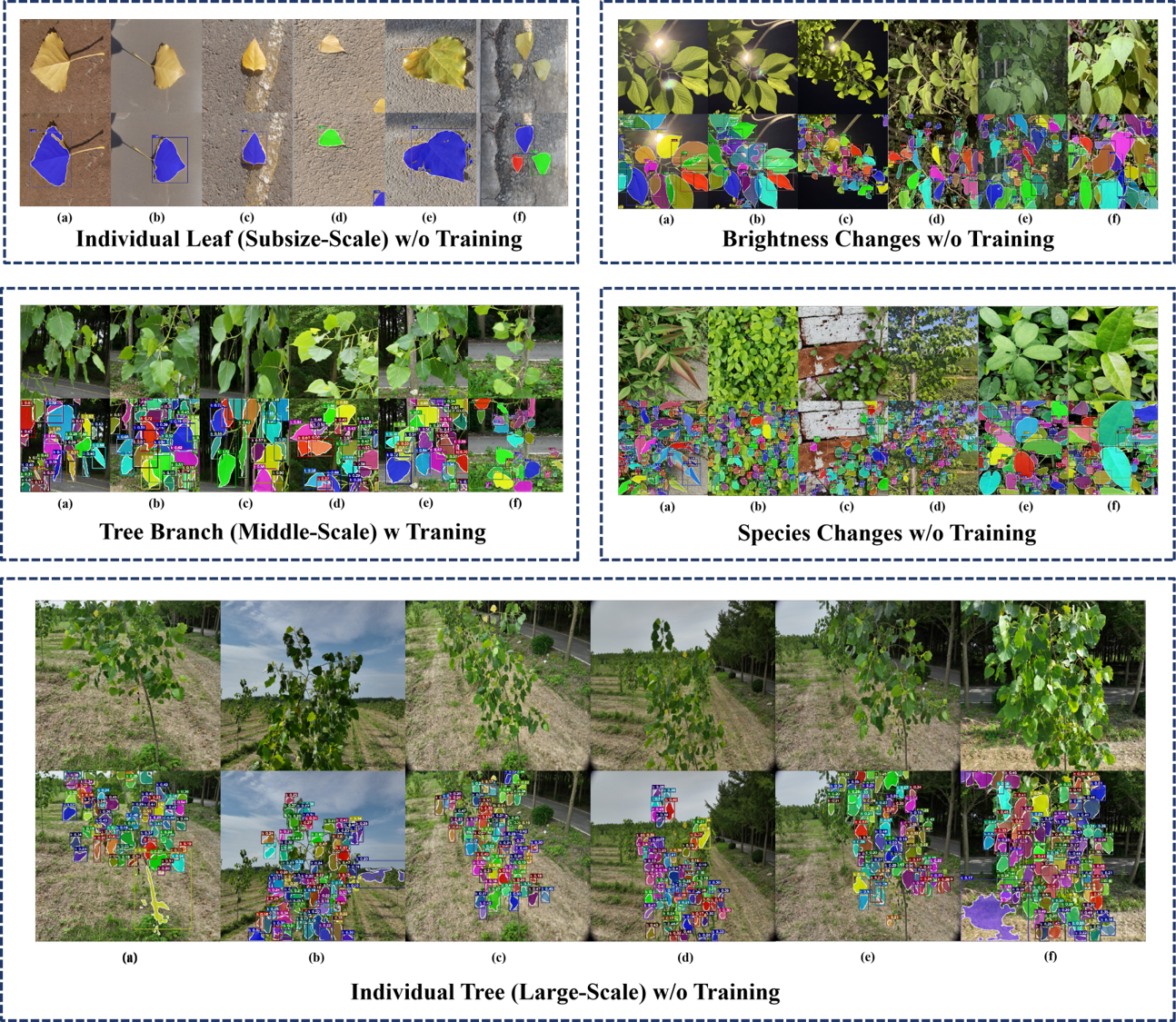}
	\caption{Segmentation results of different conditions by LeafInst.}
	\label{fig:14}
\end{figure*}
\begin{figure*}[t]
	\centering
	\includegraphics[width=0.8\textwidth]{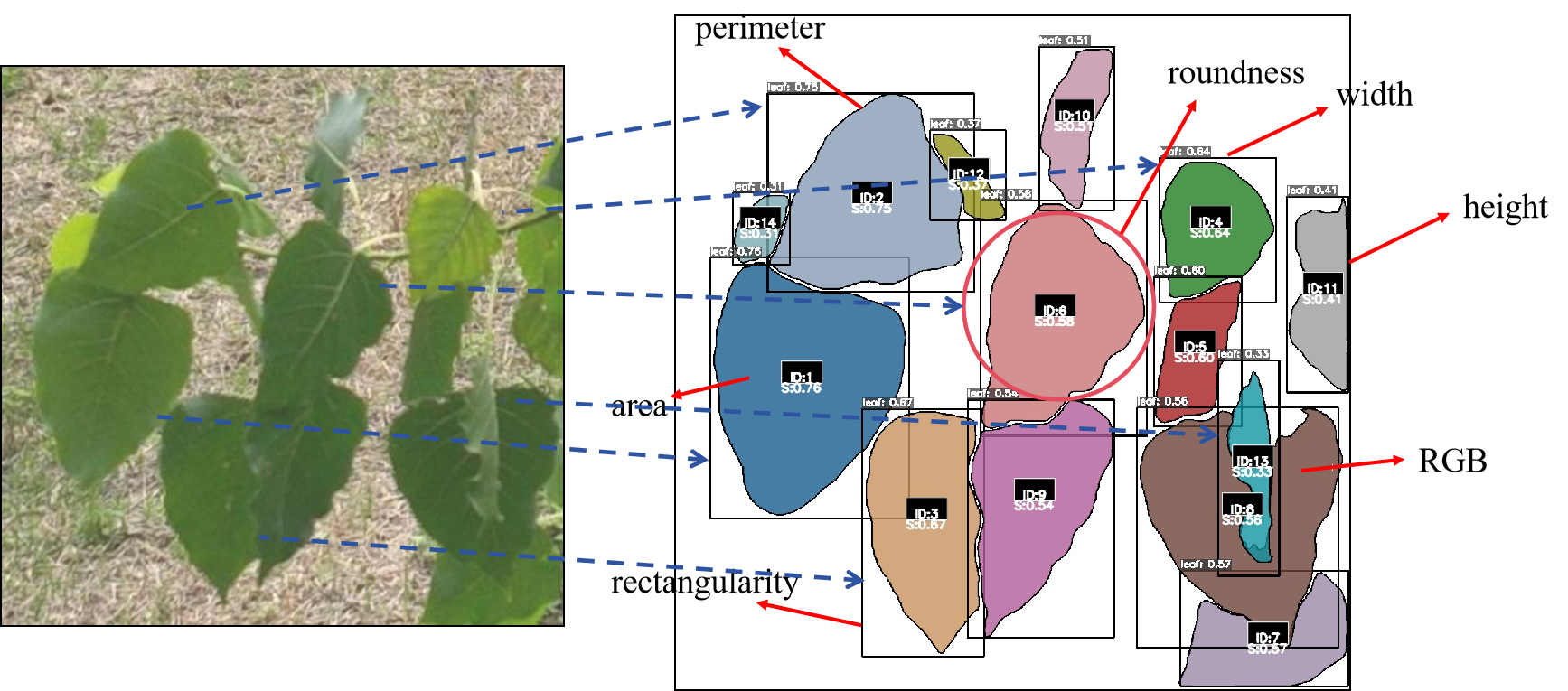}
	\caption{Schematic diagram of phenotypic indicators - taking Leaf\#306 as a example.}
	\label{fig:20}
\end{figure*}
\section{Results}
In this experiment, we used both quantitative and qualitative methods to analyze our results. Our analysis covered most of the common methods in the field of instance segmentation in recent years, namely: QueryInst \cite{63query}, Cascade-Mask R-CNN, CondInst, Mask R-CNN, SOLO, SOLOv2, PointRend \cite{62pt}, BoxInst \cite{64box}, YOLOv8-Seg, YOLOv11-Seg, YOLACT, Mask2Former, MaskDINO \cite{65md}, SparseInst, and LeafInst. These methods encompass different branch models in the fields of object detection and instance segmentation. Additionally, we also compare to the Segment Anything(SAM) Large Vision model to validate the performence of our method.
\subsection{Quantitative Results}
The quantitative results are divided into two parts, corresponding to experiments conducted on the Poplar-leaf dataset and the PhenoBench dataset, respectively. All models were trained under strictly identical experimental configurations. The quantitative results are summarized in Table~\ref{tab:val_results} and Table~\ref{tab:test_results}, which reports the validation and test set results on the Poplar-leaf dataset. The experiments demonstrate that the proposed LeafInst model achieves leading performance across the vast majority of evaluation metrics. On the Poplar-leaf validation set, LeafInst achieves a seg-mAP score of 68.4, indicating that the model is capable of capturing more fine-grained feature information and producing the most accurate segmentation results under comprehensive scenarios. As the IoU threshold increases, for example, the seg-AP50 reaches 88.9, which also ranks first among all compared models, slightly surpassing YOLOv11-Seg at 88.1. In terms of Box metrics, our model remains highly competitive and even outperforms frameworks specialized for object detection. By introducing the DARH head at the object detection stage and increasing the loss weight of the Box branch, the model significantly improves bounding box regression performance while maintaining stable segmentation accuracy. Compared with the latest Transformer-based architectures, LeafInst still shows a clear advantage. Notably, MaskDINO achieves a seg-mAP of 65.3 (ranked second) and a box-mAP of 61.1, both of which are lower than LeafInst’s 68.4 and 65.6, respectively, demonstrating the potential of our architecture. In the test set comparison, LeafInst continues to achieve the highest overall scores. It is worth noting that the test set primarily consists of large-scale leaf samples captured at close range, which places greater emphasis on cross-scale inference ability than the validation set. The results show that LeafInst ranks first across all five evaluation metrics. In particular, its seg-mAP reaches 70.0, which is 5.1\% higher than the second-best method, SoloV2 (64.9). Although its box performance under IoU = 50 is slightly lower than that of YOLOv8 (-0.01), its average box-mAP is higher (65.8).\par
On the PhenoBench dataset, we select the most competitive models for comparison, and the quantitative results are reported in Table~\ref{tab:phenobench}. The results show that LeafInst achieves the highest box-mAP of 52.7, significantly outperforming the remaining models and exceeding MaskDINO by 3.4 points (49.3). Meanwhile, LeafInst also ranks first in box-AP50 with a score of 82.3, further demonstrating its strong localization accuracy under IoU=50. For segmentation metrics, although LeafInst’s seg-mAP (50.2) is slightly lower than that of MaskDINO (51.0, -0.8), it still remains among the top-tier models. In addition, LeafInst achieves a seg-AP50 score of 80.5, slightly surpassing MaskDINO’s 80.3 (+0.2). These two datasets represent forestry UAV leaf damage scenarios and agricultural orthophoto remote sensing scenarios, respectively. Together, they further demonstrate that under complex and challenging conditions, LeafInst consistently outperforms state-of-the-art methods in both instance segmentation and object detection tasks overall.
\begin{table*}[t]
	\centering
	\caption{Manual visual inspection accuracy (\%) under different evaluation settings.}
	\label{tab:vis-acc}
	\begin{tabularx}{\textwidth}{X *{5}{C}}
		\toprule
		Method & F1-Acc & F2-Acc & F3-Acc & F4-Acc & F5-Acc \\
		\midrule
		LeafInst     & 90.9 & 86.6 & 63.1 & 88.8 & 66.8 \\
		SAM          & 81.8 & 28.8 & 18.4 & 22.2 & --   \\
		FastSAM      & 63.6 & 78.5 & 32.0 & 72.2 & 61.1 \\
		semantic-SAM & 86.3 & 67.3 & 56.3 & 87.5 & 71.0 \\
		\bottomrule
	\end{tabularx}
\end{table*}

\begin{table*}[h]
	\centering
	\caption{Ablation study of different modules.}
	\label{tab:4}
	\begin{tabular}{cccccccccc}
		\hline
		Size & Baseline & AFPN & DSAP(w/o) & DARH & TAFU & TCFU & Weight & seg/mAP & box/mAP \\ \hline
		512  & \checkmark        &      &           &      & \checkmark    &      &        & 64.5     & 59.6     \\
		1024 & \checkmark        &      &           &      & \checkmark    &      &        & 65.3     & 60.2     \\
		1024 & \checkmark        & \checkmark    &           &      & \checkmark    &      &        & 65.6     & 61.7     \\
		1024 & \checkmark        & \checkmark    & \checkmark         &      & \checkmark    &      &        & 64.5     & 60.6     \\
		1024 & \checkmark        & \checkmark    & \checkmark         & \checkmark    & \checkmark    &      & \checkmark       & 67.7     & 64.6     \\
		1024 & \checkmark        & \checkmark    & \checkmark         & \checkmark    &      & \checkmark    & \checkmark      & 68.4     & 65.8     \\ \hline
	\end{tabular}
\end{table*}
\subsection{Qualitative Results}
As shown in Figure \ref{fig:10}, the visualization results of all models on the Poplar-leaf validation set are presented, with  prediction results listed from left to right. Blue and yellow bounding boxes in the figures are used to zoom in on details for clearer elaboration. We selected images under different leaf morphologies and lighting conditions for display. In the visualization scheme, each instance is drawn in red, accompanied by a black bounding box to show the box prediction results, which also helps researchers compare models under the same pattern more effectively. The results indicate that our model's drawn leaf boundaries are more consistent with the label annotations in the GT maps for identifying various abnormal leaves, and it can even predict some boundaries overlooked by humans. Meanwhile, some networks such as MaskDino, Mask2former fall into visual hallucinations, which result it some duplicate detections. Compared to most of existing methods, LeafInst demonstrate its compatibility capability to segment poplar's leafs under different regions. This may stem from the enhanced extraction of abnormal features by the integrated DASP module, while also reflecting that the AFPN module better integrates the basic feature information of the backbone network. \par
As shown in Figure \ref{fig:11}, these are the visualization results of the model on PhenoBench. All models were trained on Poplar-leaf without specific adaptation to PhenoBench. The application to the PhenoBench dataset relies entirely on zero-shot transfer learning from our dataset and models themselves. Therefore, we used colored instance visualization results for plotting on this dataset to evaluate the robustness of different models in detecting other leaf types. White annotation boxes are used to highlight details requiring attention in the visualization results, as well as the shortcomings or advantages of each model. As indicated, our model still outperforms other baseline models in transfer capabilities for agricultural leaves across three different scenarios. Visually, our model demonstrates high precision in segmenting three different types of agricultural leaves in the PhenoBench dataset. It can accurately distinguish both narrow-leaf and broad-leaf crops. Notably, due to the angular overlap in orthographic images, many models like QueryInst often mistake different leaves from the same plant as a single entity. In contrast, methods such as MaskDino tend to miss fine details in densely packed regions during segmentation tasks. This not only demonstrates the compatibility of our model but also reflects the data support provided by our dataset for phenotypic transfer across various leaves, further validating the robustness of both the model and the dataset.\par

Figure~\ref{fig:12} presents a qualitative comparison of segmentation performance between our model and the SAM series under diverse scene challenges. The selected samples cover variations in scale, illumination, and leaf morphology. Table~\ref{tab:vis-acc} further reports the manual visual inspection accuracy (Acc) for quantitative comparison. Since the SAM series does not provide explicit instance-level outputs, precise IoU-based instance evaluation is not applicable. To ensure fairness, visually distinguishable leaf semantics are treated as leaf instances during manual inspection.\par
LeafInst is trained for 36 epochs on the Poplar-leaf dataset, while most evaluated scenes represent zero-shot transfer scenarios, except for the first sample, which is drawn from Poplar-leaf. All SAM-related models are evaluated using official pretrained weights released by their authors~\cite{66Sa,67Fastsam,68dp,69Sam2}. As shown in Figure~12-1, under the branch-scale segmentation scenario, our model produces finer and more stable leaf instance boundaries than the SAM series, achieving an Acc of 90.9, which exceeds the second-best semantic-SAM by 4.6 points. Figure~12-2 illustrates an agricultural scene sampled from PhenoBench, where our model still demonstrates clear superiority with an Acc of 86.6, followed by FastSAM with an Acc of 78.5. Figure~12-3 corresponds to an upward-looking nighttime scene under moonlight illumination, where all models exhibit degraded performance due to occlusion and shadow confusion. In this challenging setting, only LeafInst maintains a reasonable performance, reaching an Acc of 63.1. Figure~12-4 shows a wide-leaf vegetation scene adjacent to a wall, in which semantic-SAM (87.5) and LeafInst (88.8) achieve comparable results. However, in the full-tree-scale evaluation, semantic-SAM slightly outperforms LeafInst with an Acc of 71.0. Overall, while the SAM series shows limitations in specialized leaf instance segmentation tasks, particularly under scale and illumination variations, LeafInst consistently demonstrates stronger robustness and adaptability across diverse forestry and agricultural scenes.

\subsection{Ablation Study}
This section analyzes the model's performance on the Poplar-leaf validation dataset under different parameter settings through quantitative methods. The model's baseline is CondInst. As shown in Table \ref{tab:4}, AFPN indicates whether the AFPN module replaces the FPN module; DASP (w/0) means using only two asymmetric convolution branches in the DASP module without shallow-deep convolution branches; DAGH represents the complete built-in DASP module combined with dual-residual structure. TAFU represents Top-down Pixel-wise Addition Feature Fusion. TCFU represents Top-down Concatenation - decoder Feature Fusion. Weight indicates whether to use a weight of 2.0 to adjust the Box and Mask losses while reducing the centerness loss weight to 0.5. The results show that after replacing FPN with AFPN, the detection performance significantly improves, with box mAP reaching 61.7 (+1.5). This improvement stems from AFPN’s progressive multi-scale feature aggregation, which facilitates effective information exchange across different pyramid levels and enhances the representation of small-scale features, thereby improving the accuracy of the segmentation module (seg/AP: 65.6, +0.3). When only fusing DASP (w/o), the model's performance is disturbed to some extent, indicating that using only asymmetric convolution branches without integrating normal convolution branches leads to the loss of key information. After forming the complete DAGH module, this issue is resolved, and the detection performance significantly improves to 64.6 (+4.0). Meanwhile, improvements in the detection head further benefit the segmentation head, resulting in a substantial increase in segmentation performance, with the seg-AP reaching 67.7 (+3.2). After replacing TCFU with TAFU, both metrics reached their highest values, with seg-mAP at 68.4 and box-mAP at 65.8. \par
Figure \ref{fig:13} shows the iteration status during our model's training. We trained for a total of 8,064 iterations with a batch size of 4, equivalent to 36 epochs. The model fitted the dataset at an extremely fast rate, with a stable decline in loss throughout training. By adjusting the weight strategy, we ensured consistent loss magnitudes during the search for the optimal solution. The metrics generally trended positively and stabilized after approximately 4200 iterations, indicating that the model had converged.\par

\begin{table*}[h]
	\centering
	\caption{Prediction (pred) and Ground Truth (gt) Indicators - taking Leaf\#306 as a example}
	\begin{tabular}{|c|c|c|c|c|c|c|c|c|c|c|c|c|}
		\hline
		\multicolumn{13}{|c|}{Prediction (Pred)} \\
		\hline
		ID & width & height & perimeter & area & round & rect & $R_m$ & $G_m$ & $B_m$ & $\bar{R}$ & $\bar{G}$ & $\bar{B}$\\
		\hline
		1 & 294.00 & 386.00 & 1135.99 & 78915.50 & 0.77 & 0.74 & 43.00 & 71.00 & 12.00 & 44.18 & 71.65 & 13.99 \\
		2 & 296.00 & 299.00 & 1008.25 & 60085.50 & 0.74 & 0.69 & 44.00 & 75.00 & 10.00 & 46.94 & 76.24 & 12.38 \\
		3 & 176.00 & 370.00 & 964.91 & 48527.50 & 0.65 & 0.75 & 32.00 & 62.00 & 6.00 & 34.16 & 62.99 & 8.81 \\
		4 & 173.00 & 208.00 & 640.84 & 26074.00 & 0.80 & 0.75 & 79.00 & 104.00 & 12.00 & 80.73 & 104.55 & 12.93 \\
		5 & 132.00 & 218.00 & 611.06 & 17917.50 & 0.60 & 0.77 & 37.00 & 64.00 & 11.00 & 40.81 & 65.70 & 13.78 \\
		6 & 249.00 & 343.00 & 1002.59 & 56927.50 & 0.71 & 0.69 & 30.00 & 58.00 & 11.00 & 32.77 & 59.66 & 12.34 \\
		7 & 258.00 & 175.00 & 856.48 & 26471.50 & 0.45 & 0.60 & 41.00 & 73.00 & 8.00 & 42.13 & 73.67 & 9.74 \\
		8 & 310.00 & 369.00 & 1602.36 & 65113.00 & 0.32 & 0.59 & 23.00 & 51.00 & 5.00 & 26.06 & 52.05 & 7.06 \\
		10 & 112.00 & 244.00 & 613.24 & 15567.00 & 0.52 & 0.72 & 91.00 & 105.00 & 76.00 & 88.42 & 105.04 & 77.11 \\
		12 & 111.00 & 132.00 & 394.72 & 6371.50 & 0.51 & 0.70 & 83.00 & 106.00 & 46.00 & 87.97 & 109.40 & 50.74 \\
		13 & 71.00 & 292.00 & 682.62 & 12598.00 & 0.34 & 0.65 & 58.00 & 79.00 & 36.00 & 59.08 & 79.47 & 37.36 \\
		14 & 85.00 & 104.00 & 316.59 & 5903.00 & 0.74 & 0.74 & 124.00 & 153.00 & 9.00 & 122.55 & 150.32 & 13.82 \\
		\hline
	\end{tabular}
	\begin{tabular}{|c|c|c|c|c|c|c|c|c|c|c|c|c|}
		\hline
		\multicolumn{13}{|c|}{Ground Truth (GT)} \\
		\hline
		ID & width & height & perimeter & area & round & rect & $R_m$ & $G_m$ & $B_m$ & $\bar{R}$ & $\bar{G}$ & $\bar{B}$\\
		\hline
		1 & 289.00 & 392.00 & 1158.51 & 78377.00 & 0.73 & 0.73 & 43.00 & 71.00 & 12.00 & 44.01 & 71.52 & 13.80 \\
		2 & 289.00 & 293.00 & 976.00 & 58223.00 & 0.77 & 0.72 & 44.00 & 75.00 & 9.00 & 46.55 & 76.03 & 12.07 \\
		3 & 172.00 & 374.00 & 977.44 & 47354.00 & 0.62 & 0.75 & 32.00 & 62.00 & 6.00 & 33.85 & 62.79 & 8.37 \\
		4 & 174.00 & 213.00 & 662.11 & 27172.50 & 0.78 & 0.77 & 79.00 & 104.00 & 12.00 & 80.83 & 104.31 & 13.67 \\
		5 & 133.00 & 216.00 & 615.50 & 17864.00 & 0.59 & 0.78 & 37.00 & 64.00 & 11.00 & 40.74 & 65.75 & 13.60 \\
		6 & 250.00 & 357.00 & 1051.99 & 56756.50 & 0.64 & 0.66 & 30.00 & 58.00 & 11.00 & 32.78 & 59.69 & 12.34 \\
		7 & 258.00 & 177.00 & 904.83 & 26094.50 & 0.40 & 0.58 & 41.00 & 73.00 & 8.00 & 42.66 & 74.29 & 9.97 \\
		8 & 313.00 & 381.00 & 1714.04 & 63251.50 & 0.27 & 0.54 & 23.00 & 50.00 & 5.00 & 25.31 & 51.41 & 6.54 \\
		10 & 111.00 & 249.00 & 625.29 & 15393.50 & 0.49 & 0.73 & 91.00 & 104.00 & 75.00 & 88.47 & 105.05 & 76.91 \\
		12 & 105.00 & 108.00 & 357.75 & 5157.50 & 0.51 & 0.70 & 79.00 & 104.00 & 40.00 & 84.82 & 108.15 & 45.07 \\
		13 & 69.00 & 325.00 & 789.97 & 14555.50 & 0.29 & 0.70 & 56.00 & 77.00 & 35.00 & 57.65 & 78.19 & 36.09 \\
		14 & 90.00 & 110.00 & 333.32 & 6265.50 & 0.71 & 0.74 & 123.00 & 153.00 & 10.00 & 120.14 & 147.85 & 14.12 \\
		\hline
	\end{tabular}
	\label{tab:306}
\end{table*}

\section{Applications}
\begin{figure*}[t]
	\centering
	\includegraphics[width=1\textwidth]{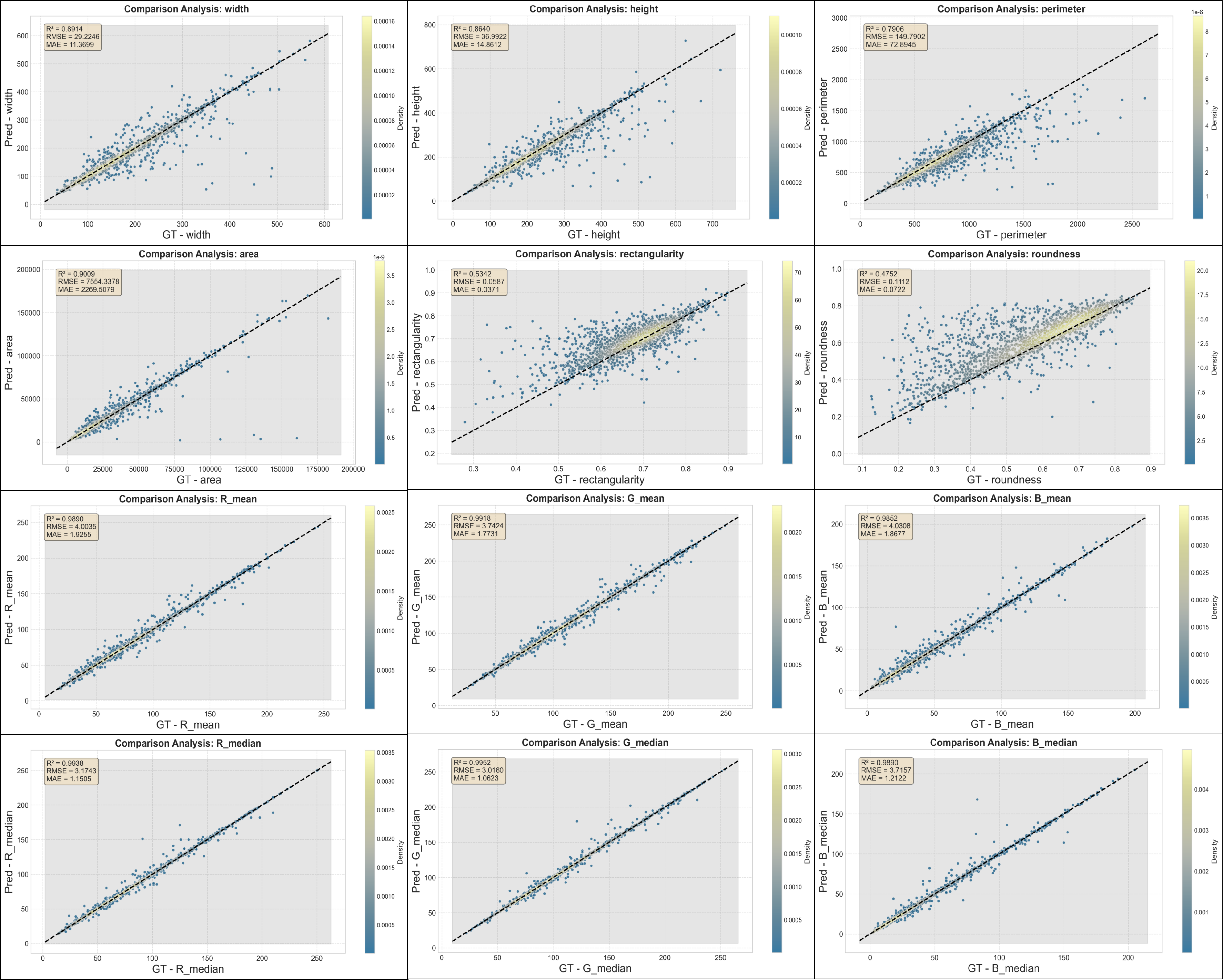}
	\caption{Regression curves of predicted and GT secondary Indicators.}
	\label{fig:15}
\end{figure*}

To comprehensively demonstrate the value of our model in intelligent forestry breeding, we conducted extensive application experiments to verify the robustness of the model and data, and provided new possibilities and directions for the integration. We conduct our experiments on two downstream application tasks. The first task is the application of leaf segmentation and counting in different scenarios. The second branch is to use LeafInst to establish a regression indicator for calculating leaf growth and vegetation greenness for "early-stage" selection.

\subsection{Leaf Instances Segmentation Under Different Conditions}
In this subsection, we elaborate on how LeafInst is used for leaf segmentation to verify leaf counting performance, with multiple different control group environments set up. In terms of scale, the scenarios are divided into single-leaf scenes, branch scenes, and whole-tree scenes. In terms of environment, we compare daytime and nighttime lighting conditions. In terms of leaf species, we introduce leaves of different species such as tea trees, Nandina domestica, and Trachelospermum jasminoides. These scenarios cover common conditions and hold forestry significance. Importantly, these new scenarios are basically untrained with specific data, relying entirely on the parameters learned by LeafInst from Poplar-leaf for zero-shot transfer learning. The results are shown in Figure\ref{fig:14}.\par

In the single-leaf scenario, the model completed the leaf segmentation work with pixel-level accuracy. It is worth noting that the samples we selected are deciduous leaves, which are yellow in color. However, this still does not affect the model's ability to recognize the leaf contours, further proving that LeafInst has learned the color and shape features.\par

In the branch scenario, the model successfully identified deformed leaves of different shapes by virtue of the asymmetric convolution of the DASP module. In the face of situations such as uneven lighting and overlapping occlusions, LeafInst can identify the position and range of most leaves. It should be noted that sometimes some background leaves are misidentified. These errors can be completely filtered out by selecting target boxes with a lower threshold or deleting instances with a too-low pixel range, and they do not affect the application performance.\par
\begin{table*}[t]
	\centering
	\caption{The optimal value of intermediate indicators of colors}
	\label{tab:opt}
	\begin{tabular}{cccccccccccc}
		\toprule
		$R_m$ & $\bar{R}$ & $R_{1/3}$ & $R_{2/3}$ & $G_m$ & $\bar{G}$ & $G_{1/3}$ & $G_{2/3}$ & $B_m$ & $\bar{B}$ & $B_{1/3}$ & $B_{2/3}$ \\
		\midrule
		80.02 & 82.50 & 73.71 & 87.15 & 111.68 & 113.10 & 105.49 & 118.38 & 41.39 & 44.45 & 34.40 & 49.36 \\
		\bottomrule
	\end{tabular}
\end{table*}

In the individual-tree scenario, the trees show various postures, and the branches are obviously offset due to the influence of the wind. Moreover, due to the huge sampling scale, the area occupied by a single leaf instance in the picture is extremely small. This is much smaller than the training samples in Poplar-leaf. Even in this highly dense situation, LeafInst still captures the leaf information of young poplar trees with high accuracy. Although there are slight errors in some areas, for example, the yellow branch in the lower left corner of the first column is misidentified. But this is because we downsampled the original image to 1024 $\times$ 1024, resulting in the loss of a lot of information, making the image not high-definition. Maintaining the original resolution will alleviate these problems. In short, in the scenario of scale transformation, the model proves its potential capability. \par
In the nighttime scenario, the lighting is insufficient, and the background is completely different from the well-lit scenarios in Poplar-leaf, which poses a challenge for the model to recognize shadows. The results show that without being trained specifically for shadows, LeafInst has some errors in shadow recognition, that is, it mistakenly identifies shadows as overlapping leaves. Nevertheless, the model has overcome some of the challenges brought by shadows, which is acceptable in the case of zero-shot training, and it also provides new application possibilities for real-time monitoring of leaf characteristics at night. \par
Under the challenge of leaf scenarios of different tree species which present dfferent shape, our model demonstrates adaptability and transferability. For example, the Nandina domestica in the picture has sharp-shaped leaves, the leaves of vine plants are polygonal, the tea plants have a large number of dense leaves, and the leaves of Trachelospermum jasminoides are slender. These leaf shapes are completely different from the broad-leaved genus of poplar leaves, but LeafInst still shows acceptable recognition accuracy. \par
\begin{table*}[htbp]
	\centering
	\caption{Statistical table of intermediate indicators. $R_m$ denotes the median value, $\bar{R}$ denotes the mean value, $R_{1/3}$ denotes the lower third quartile, $R_{2/3}$ denotes the higher third quartile. Other channels are same as R channel.}
	\label{tab:sta}
	\small
	\begin{tabular}{lcccccc}
		\toprule
		\multicolumn{7}{c}{Shape Indicators} \\
		\midrule
		& width & height & perimeter & area & roundness & rectangularity \\
		\midrule
		$R^2$  & 0.891 & 0.864 & 0.791 & 0.901 & 0.475 & 0.534 \\
		RMSE    & 29.225 & 36.992 & 149.790 & 7554.338 & 0.111 & 0.059 \\
		MAE     & 11.370 & 14.861 & 72.894 & 2269.508 & 0.072 & 0.037 \\
		\bottomrule
	\end{tabular}
	\vspace{0.5em}
	\begin{tabular}{lcccccccccccc}
		\toprule
		\multicolumn{13}{c}{Color Indicators} \\
		\midrule
		& $R_m$ & $G_m$ & $B_m$ & $\bar{R}$ & $\bar{G}$ & $\bar{B}$ & $R_{1/3}$ & $G_{1/3}$ & $B_{1/3}$ & $R_{2/3}$ & $G_{2/3}$ & $B_{2/3}$ \\
		\midrule
		$R^2$ & 0.994 & 0.995 & 0.989 & 0.989 & 0.992 & 0.985 & 0.989 & 0.991 & 0.986 & 0.992 & 0.994 & 0.987 \\
		RMSE   & 3.174 & 3.016 & 3.716 & 4.004 & 3.742 & 4.031 & 4.035 & 4.023 & 3.777 & 3.673 & 3.386 & 4.502 \\
		MAE    & 1.151 & 1.062 & 1.212 & 1.925 & 1.773 & 1.868 & 1.121 & 1.169 & 1.137 & 1.504 & 1.307 & 1.541 \\
		\bottomrule
	\end{tabular}
\end{table*}

\begin{table*}[t]
	\centering
	\caption{Indicator weights of LGCI (keep two decimals) }
	\begin{tabularx}{1\textwidth}{*{6}{>{\centering\arraybackslash}X}}
		\toprule
		\multicolumn{6}{c}{Shape Indicators Weights} \\
		\midrule
		width & height & perimeter & area & roundness & rectangularity \\
		\midrule
		0.15 & 0.11 & 0.13 & 0.30 & 0.06 & 0.03 \\
		\bottomrule
	\end{tabularx}
	\begin{tabularx}{1\textwidth}{*{12}{>{\centering\arraybackslash}X}}
		\toprule
		\multicolumn{12}{c}{Color Indicators Weights} \\
		\midrule
		$R_m$ & $G_m$ & $B_m$ & $\bar{R}$ & $\bar{G}$ & $\bar{B}$ & $R_{1/3}$ & $G_{1/3}$ & $B_{1/3}$ & $R_{2/3}$ & $G_{2/3}$ & $B_{2/3}$ \\
		\midrule
		0.02 & 0.03 & 0.01 & 0.02 & 0.03 & 0.01 & 0.01 & 0.02 & 0.01 & 0.02 & 0.03 & 0.02 \\
		\bottomrule
	\end{tabularx}
	\label{tab:wei}
\end{table*}
\begin{figure*}[t]
	\centering
	\includegraphics[width=1\textwidth]{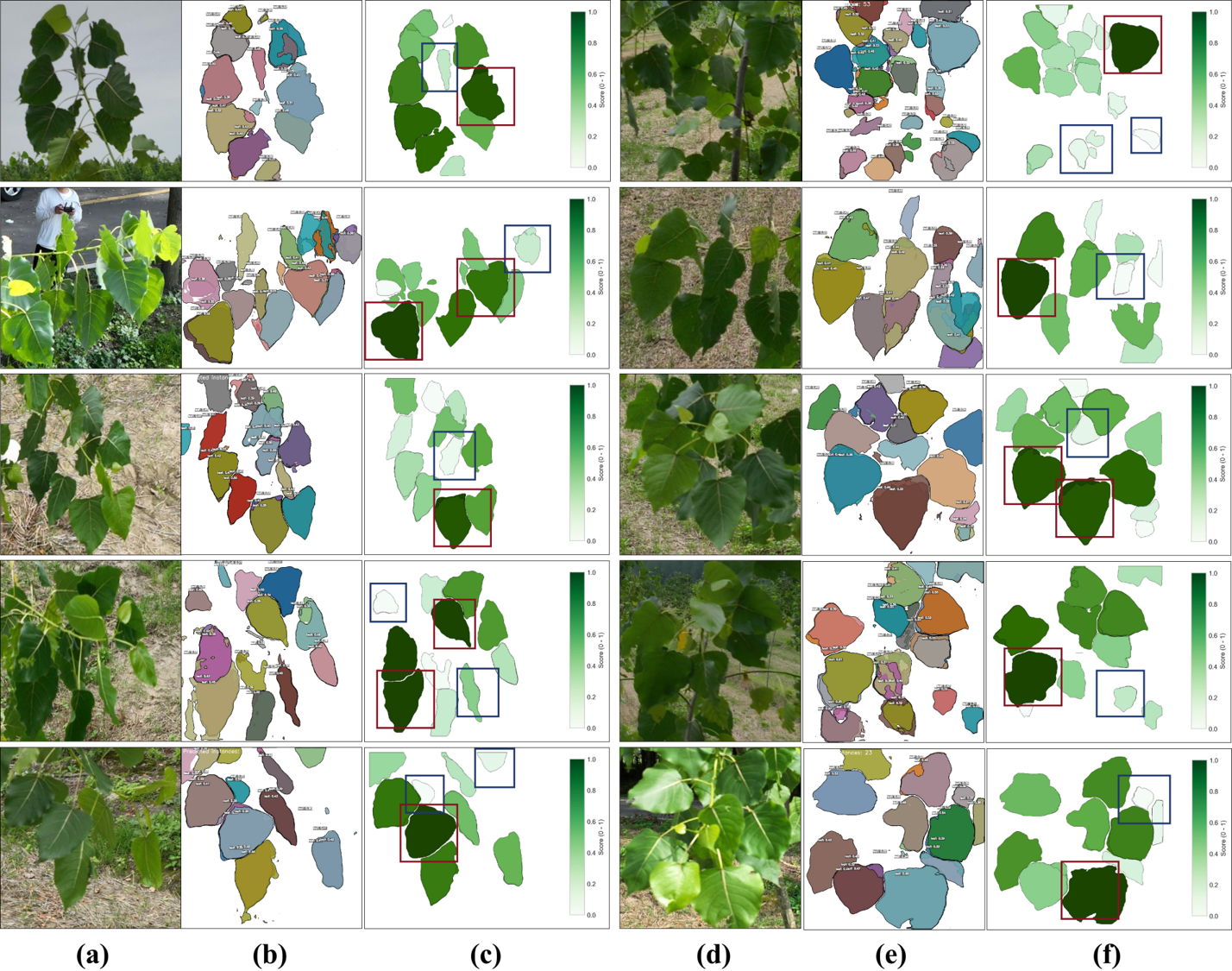}
	\caption{Quantitative results of the Leaf Growth Condition Indicator (LGCI) scores. (a, d) Input images; (b, e) instance segmentation results produced by LeafInst (IoU > 0.3); (c, f) LGCI score visualization maps. Red and blue boxes highlight leaf instances exhibiting relatively favorable (positive) and unfavorable (negative) morphological characteristics, respectively}
	\label{fig:17}
\end{figure*}
\subsection{Leaf Growth Condition Indicator}
In this section, we will present a novel UAV-RGB image based indicator to calculate the leaf growth condition using our proposed network. We used LeafInst to identify and obtain 1,850 representative leaves from the Poplar-leaf validation set as the research data foundation. Key parameters of each leaf instance were identified to evaluate leaf growth. These parameters are vegetation phenotypic indicators computed from the instance segmentation masks, with all measurements expressed in pixel units. Using these parameters combined with the entropy weight method, we proposed a new leaf growth condition parameter based on UAV images: \textit{Leaf Growth Condition Indicator} (LGCI). Subsequent experiments suggest that this indicator has the potential to automatically characterize leaves with desirable shape properties. The key parameters used for modeling this indicator are divided into two parts. The specific indicators calculation process can be seen in Figure \ref{fig:20}. Taking one image as an example, each instance was assigned a unique ID through matching with the original image, and its corresponding phenotypic indicators were calculated, as shown in Table \ref{tab:306}. The first part is shape parameters: length, width, perimeter, area, roundness, and rectangularity. These indicators measure leaf growth by extracting leaf shapes. Since the spatial variation in a single UAV image is not significant, these indicators are all positive indicators after normalization to eliminate the influence of dimensions. The formulas for roundness and rectangularity are as follows:
\begin{align}
	&& \mathrm{roundness} = \frac{4 \cdot \pi \cdot A}{C^{2}}, &\\
	&& \mathrm{rectangularity} = \frac{A}{(x_{\max}-x_{\min}) \cdot (y_{\max}-y_{\min})}, &
\end{align}
where $C$ denotes the perimeter of a leaf, $A$ denotes the leaf area, and 
$x_{\min}, x_{\max}, y_{\min}, y_{\max}$ represent the coordinates of the minimum bounding rectangle enclosing the leaf.

The second part of the indicators focuses on the color information of the leaves, including the mean, median, and tertiles of each RGB channel. These colors imply internal structural information of the leaves, such as reflectivity, moisture content, and lesion status of each band. These indicators are intermediate indicators that need to be normalized through statistical analysis. The current feasible approach is to convert color features from intermediate indicators to positive indicators through an optimal value table. This article focuses on the collected data, and invites forestry experts to select high-quality leaves and conduct RGB information big data statistics on the high-quality leaves to obtain the optimal value table of these indicators as shown in Table \ref{tab:opt}. Then, we transform the color features from intermediate-type indicators into positive indicators using it. \par

Next, use LeafInst to predict these indicators. By matching and aligning the predicted instances with the annotated GT one by one, a prediction error map can be obtained. Figure \ref{fig:15} shows the regression curves of the predicted values and GT values of each indicator. Table \ref{tab:sta} shows the stastical value of each indicator. It can be seen that our model has high prediction accuracy for most indicators. Most secondary indicators keep high $R^{2}$ score, which are on average of 0.9. Although some indicators such as $roundness$ and $rectangularity$ present fluctuations. However, the absolute error of them such as MAE are 0.072, 0.037 respectively, which are low enough to accept this due to the instability inhernet in the calculation of these indicators themselves. The data ensures the accuracy of phenotypic data extraction and provides automatic extraction technical support for subsequent applications. Regarding the alignment strategy, we match the predictions with the GT based on a threshold of the lowest IoU of 50 to ensure no overlap and repetition. Meanwhile, to ensure that low - quality leaves are not recorded, we filter out instances with an area of less than 1000 pixels. Finally, we obtained the weight values of LGCI through the entropy weight method as shwon in Table \ref{tab:wei}. This indicator represents the first leaf growth status scoring coefficient derived from UAV-based RGB imagery. Using this indicator, we demonstrate the feasibility of assessing leaf growth status using RGB information alone, without relying on additional data such as multispectral bands. It can be seen that the shape indicator has a relatively large weight, indicating that among the indicators used to evaluate the growth status of leaves, shape information is of vital importance. Although the secondary indicators of color band information seem to have a relatively low individual weight, when combined, they still play a crucial role in the calculation of the final LGCI. In this way, it is ensured that the calculation of LGCI implicitly contains information about the shape and internal reflectance.  \par

As shown in Figure \ref{fig:17}, LeafInst uses the LGCI calculation to visualize and judge the growth status of leaf phenotypic instances. We calculated the indicators for all predicted instances, mapped them to the range of 0-1, and conducted a qualitative analysis of each instance to evaluate the performance of the indicators. A darker shade reflects a better growth status of the plant leaves. Among them, the red rectangular boxes in the figure are used to highlight the recognition results of the positive examples of the model, and the blue rectangular boxes are used to emphasize the results of the negative examples of the model. From the instance prediction diagram to the LGCI visualization diagram, it can be seen that some instances have been screened and deleted due to factors such as small area or IoU overlap. The results show that broad, plump, and fresh green leaves received a higher LGCI score, while slender and broken leaves received a lower LGCI score. This demonstrates the application value of this indicator. The high-throughput characteristic of this indicator cannot be achieved by human field surveys. Field surveys are labor-intensive and may involve destructive sampling procedures, which can potentially damage parent plants and increase the risk of contamination. However, by combining this indicator with an instance segmentation model, it is possible to quickly and on a large scale select plants with excellent shapes for cultivation, which has great value in forestry. It provides a technical means and quantitative support for early-stage selection and breeding, which is one of the bottleneck problems in the intelligent breeding of plantation forests at the current stage. \par

\section{Conclusions}
In this study, we addressed the long-standing challenge of fine-grained forestry leaf phenotypic analysis in open-field environments by proposing a unified instance segmentation framework, LeafInst, together with a high-quality UAV-based dataset. The Poplar-leaf dataset is publicly released as the first pixel-level annotated instance segmentation dataset specifically designed for forestry leaves in natural scenes, capturing substantial variations in scale, illumination, and leaf morphology under real growth conditions. To effectively handle these challenges, LeafInst integrates AFPN to enhance progressive multi-scale feature representation, introduces a Dynamic Anomalous Regression Head with Dynamic Asymmetric Spatial Perception to better model irregular and deformed leaf structures, and employs a Top-down Concatenation–decoder Feature Fusion strategy to alleviate feature redundancy and improve mask decoding quality. Extensive quantitative and qualitative experiments demonstrate that LeafInst consistently outperforms state-of-the-art instance segmentation methods on the Poplar-leaf dataset and shows strong zero-shot transferability on the agricultural PhenoBench dataset, indicating robust generalization across different vegetation scenarios. Furthermore, we demonstrate the practical value of instance-level leaf segmentation by proposing a high-throughput \textit{LGCI}, which enables quantitative assessment of leaf phenotypic development based on multiple secondary traits. These results indicate that the proposed dataset and framework effectively bridge the gap between instance segmentation and fine-grained leaf phenotype analysis in open-field forestry scenes, providing a scalable and reliable technical foundation for intelligent forestry breeding and downstream ecological applications.
\section{Acknownledgement}
This work was supported by Biological Breeding-National Science and Technology Major Project (NO.2023ZD0405605). 
\bibliographystyle{elsarticle-num}
\bibliography{references}

\end{document}